\PassOptionsToPackage{table}{xcolor}
\documentclass{article}

\usepackage{iclr2027_conference}
\usepackage{times}
\usepackage[utf8]{inputenc}
\usepackage[T1]{fontenc}
\usepackage{hyperref}
\usepackage{url}
\usepackage{booktabs}
\usepackage{amsfonts}
\usepackage{nicefrac}
\usepackage{microtype}
\usepackage[table]{xcolor}
\usepackage{colortbl}
\usepackage{amsmath}
\usepackage{amssymb}
\usepackage{mathtools}
\usepackage{amsthm}
\usepackage{algorithm}
\usepackage{algpseudocode}
\usepackage{paralist}
\usepackage{placeins}
\usepackage{mdframed}
\usepackage[most]{tcolorbox}
\definecolor{figdata}{HTML}{D2691E}
\definecolor{figgen}{HTML}{2A8C82}
\usepackage{subcaption}
\usepackage{wrapfig}
\usepackage{makecell}
\usepackage{multirow}
\definecolor{sftgreen}{HTML}{E8F5E9}
\definecolor{rlblue}{HTML}{E3F2FD}
\usepackage{makecell}

\newcommand{\vx}{\mathbf{x}}
\newcommand{\vy}{\mathbf{y}}

\newcommand{\methodname}{Ambient SFT}
\newcommand{\tmname}{Temperature Matching}

\usepackage[capitalize,noabbrev]{cleveref}
\usepackage{float}

\theoremstyle{plain}
\newtheorem{theorem}{Theorem}[section]
\newtheorem{proposition}[theorem]{Proposition}
\newtheorem{lemma}[theorem]{Lemma}
\newtheorem{corollary}[theorem]{Corollary}
\theoremstyle{definition}

\theoremstyle{remark}
\newtheorem{remark}[theorem]{Remark}

\title{Ambient LLMs:\\ Data-Efficient Supervised Fine-tuning via Temperature Matching}

\author{%
  Giannis Daras \\
  MIT \\
  \texttt{gdaras@mit.edu} \\
  \And
  Marios Papachristou \\
  Arizona State University \\
  \texttt{mpapachr@asu.edu}
  \And
  Adam Klivans \\
  University of Texas at Austin \\
  \texttt{klivans@cs.utexas.edu}
  \And
  Antonio Torralba \\
  MIT \\
  \texttt{torralba@mit.edu}
  \And
  Constantinos Daskalakis \\
  MIT \\
  \texttt{costis@mit.edu}
}

\iclrfinalcopy

\begin{document}

\addtocontents{toc}{\protect\setcounter{tocdepth}{-1}}

\maketitle

\begin{abstract}
    Collecting large-scale, diverse, and high-quality post-training datasets for autoregressive language models is challenging. Consequently, Supervised Fine-Tuning (SFT) is typically applied to datasets significantly smaller than those used for pre-training, which leads to overfitting and the catastrophic forgetting of the tails of the data distribution. To prevent this collapse, we introduce \methodname, a principled framework for leveraging imperfect predictions to augment the fine-tuning set. In our method, a ``frozen'' version of the pre-trained model serves as a noisy teacher that the fine-tuned model must match at a coarse level. In particular, we require that the high-temperature versions of the conditional distributions produced by the post-training model should match the high-temperature versions of the pre-training ones. The temperature hyperparameter controls the level of trust in the pre-trained model, with the limit $T\to \infty$ reverting back to regular SFT and $T=1$ to knowledge distillation. We demonstrate that our approach outperforms standard SFT across domain adaptation, instruction-tuning, and reasoning.
\end{abstract}

\section{Introduction}
\label{sec:intro}
The development of state-of-the-art Large Language Models (LLMs) has evolved into a complex, multi-stage pipeline~\citep{liu2024deepseek, bai2023qwen, abdin2024phi, achiam2023gpt, olmo2025olmo, grattafiori2024llama}. This process typically progresses from massive-scale pre-training to more specialized post-training phases, including instruction-tuning~\citep{ouyang2022training}, alignment~\citep{ziegler2019fine, ouyang2022training}, and reasoning~\citep{guo2025deepseek, lightman2023let}. As models advance through this pipeline, data availability decreases by orders of magnitude. While pre-training corpora often encompass trillions of tokens, later stages frequently rely on datasets comprising merely a few thousand highly curated examples~\citep{zhou2023lima, guha2025openthoughts}. This severe bottleneck arises because post-training requires extraordinarily high-quality data, making large-scale collection prohibitively challenging~\citep{zhou2023lima}.

The stark contrast in data scale introduces undesirable artifacts. Most notably, post-trained LLMs tend to suffer from catastrophic forgetting at the tails of the pre-training data distribution, overfitting instead to the specific stylistic patterns or narrow viewpoints overrepresented in the limited post-training data~\citep{ouyang2022training, lin2024mitigating}. This phenomenon is known as the \textit{alignment tax}~\citep{ouyang2022training}. The prevailing solution to this bottleneck is brute force: allocating vast financial and computational resources to scale up the collection of post-training data. In this work, we ask a different question: are there \textit{algorithmic solutions} that can increase the data efficiency of post-training without inducing severe model collapse or overfitting?

We start our approach to this problem by performing a thought experiment. Imagine having access to an oracle LLM that can generate the high-quality post-training data distribution, but you can only observe a high-temperature, noisy version of its predictions. We show that learning from high-temperature data can match the performance of training on the true, standard-temperature distribution, provided the sample size is large enough. For a practical algorithm, we do not have a perfect high-temperature oracle, but we can effectively approximate it using a high-temperature \textit{weak} model, which is the model prior to post-training. Distributionally, any two models converge as the sampling temperature increases; thus, at a sufficiently high temperature, a readily available weak model serves as a strong proxy for the unavailable oracle. This motivates our proposed approach: during post-training, we augment our dataset using a weak teacher operating at a high temperature. We enforce that the post-trained model matches the weak teacher's distribution at this high temperature, while remaining unconstrained to refine its conditional distributions at the standard temperature ($T=1$). The temperature hyperparameter explicitly controls the level of trust placed in the pre-trained model, with the $T \to \infty$ limit gracefully reverting to standard SFT and with $T=1$ to knowledge distillation.

Our algorithm, termed \methodname{}, mitigates the overfitting typically observed when applying Supervised Fine-Tuning (SFT) to small datasets. It achieves this by adding to the standard SFT loss a second term, which we call \tmname{}, that regularizes the post-trained network to match the high-temperature conditional distributions of its pre-trained predecessor. We evaluate \methodname{} across three standard post-training paradigms: \textit{(i)} domain adaptation, \textit{(ii)} instruction-tuning, and \textit{(iii)} the development of reasoning capabilities. Across all settings, our approach consistently outperforms standard SFT. In summary, our contributions are as follows:

\begin{compactitem}
    \item We introduce \textit{\methodname}, a drop-in replacement for standard SFT that regularizes fine-tuning by matching a high-temperature version of the model's own pre-trained distribution. A single temperature knob interpolates between SFT and knowledge distillation.
    \item We prove that a target distribution can be recovered from high-temperature conditionals alone, that any two models' conditionals contract as the temperature grows, and that under a log-linear model Ambient SFT strictly beats vanilla SFT and knowledge distillation.
    \item We show gains in all three post-training regimes: up to $26.7\%$ lower perplexity than the best baseline across five domain-adaptation targets; average relative gains of $8.0\%$ to $20.3\%$ on instruction tuning and wins on $5/6$ reasoning benchmarks.
\end{compactitem}

\begin{figure*}[!htp]
    \centering
    \includegraphics[width=0.9\linewidth]{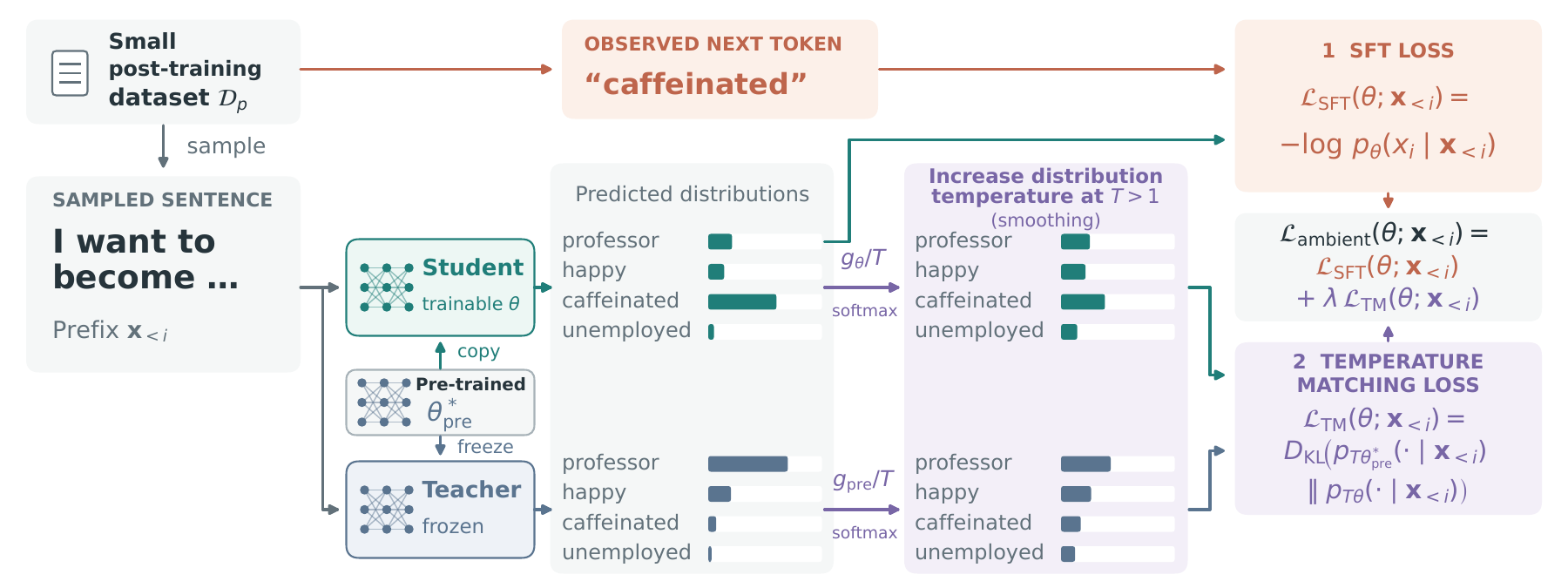}
    \caption{Overview of \methodname{} for data-efficient post-training.\small{
    A pre-trained model initializes a trainable student and a frozen teacher. On the small dataset $\mathcal D_p$, SFT fits the observed next token, while Temperature Matching aligns the two models' distributions after smoothing. Their weighted sum updates only the student. Temperature controls trust in the teacher: $T\to\infty$ recovers SFT and $T=1$ knowledge distillation.}}
    \label{fig:fig1}
\end{figure*}

\section{Background}
\noindent \textbf{Notation.}
Let $V$ be a finite alphabet of tokens, and let $L \subseteq V^{l}$ be a language associated with a probability measure $p$. A sample from that language is denoted with $\vx = (x_1, \dots, x_l)$, where $x_i \in V$. We use $\vx_{<i} = (x_1, ..., x_{i-1})$ to denote the vector with the first $(i-1)$ co-ordinates. 

\noindent \textbf{Training of Auto-Regressive (AR) Language Models.}
Any distribution, can be factorized as follows:
$p(\vx) = p(x_1) p(x_2 | x_1) \cdot ... \cdot p(x_l | \vx_{<l})$. Autoregressive Language Models (LMs) are artificial networks trained to capture these conditional distributions. Specifically, a model  $g_{\theta}(\cdot | \vx_{< i})$ 
outputs logits for each potential token in the vocabulary given a context $\vx_{<i}$, inducing conditional distributions:
\begin{gather}
    p_{\theta}(x_i \mid \vx_{<i}) =  \frac{e^{g_{\theta}(x_i | \vx_{<i})}}{\sum_{y_i \in V} e^{g_{\theta}(y_i | \vx_{<i})}}.
    \label{eq:temp_1_conditionals}
\end{gather}

The standard training objective targets the minimization of the population negative log-likelihood

\begin{equation}
    J(\theta) = \sum_{i=1}^{l} \mathbb E_{\vx_{\leq i} \sim p_{\le i}}\left[ -\log\left(\frac{e^{g_{\theta}(x_i | \vx_{<i})}}{\sum_{y_i \in V} e^{g_{\theta}(y_i | \vx_{<i})}}\right)\right],
    \label{eq:normal_loss}
\end{equation}
where $p_{\le i}$ denotes the distribution of the first $i$ tokens. Eq.~\eqref{eq:normal_loss} computes the cross-entropy between the model's predicted distribution and the target distributions of each token $x_i$ observed after its prefix $\vx_{<i}$. By minimizing this loss, the model approximates the true data conditionals $p(x_i \mid \vx_{<i})$.

\noindent \textbf{Adjusting the temperature.} At inference time, the sharpness of the learned conditional distributions can be modulated by dividing the network logits by a scalar $T > 0$, known as the \textit{sampling temperature}. We use $p_{T\theta}(x_i | \vx_{<i}) \propto p_{\theta}(x_i | \vx_{<i})^{1/T}$ to refer to the temperature-scaled distributions:
\begin{equation}
    p_{T\theta}(x_i | \vx_{<i}) = \frac{e^{g_{\theta} (x_i | \vx_{<i})/T}}{\sum_{y_i \in V} e^{g_{\theta}(y_i | \vx_{<i})/T}}.
    \label{eq:temp_scaled_conditionals}
\end{equation}

\noindent \textbf{The need for post-training.}
In practice, we do not have oracle access to the true distribution $p$. Instead, we optimize the objective of Eq.~\eqref{eq:normal_loss} using a finite empirical dataset $\mathcal{D}_p = \{ \vx^{(k)}\}_{k=1}^{n_1}$ sampled from $p$. We denote this empirical distribution with $\hat p_{n_1}$. Typically, the success of the generative training depends on the size of $n_1$. However, collecting a sufficiently large dataset directly from $p$ is often prohibitively expensive. For example, $p$ might represent the distribution of expert-level textbooks or highly specialized instruction-following dialogues. While such in-distribution data is scarce, there exists an abundance of generic text on the web, drawn from a broader, different distribution $q$. This necessitates the use of two distinct datasets: a massive dataset $\mathcal{D}_q = \{ \vy^{(k)}\}_{k=1}^{n_2}$ sampled from $q$, and our limited target dataset $\mathcal{D}_p$, where $n_1 \ll n_2$. The prevailing learning paradigm leverages both datasets by first pre-training the model on $\mathcal{D}_q$ and subsequently fine-tuning it on $\mathcal{D}_p$. Hereafter, we use $\theta^*_{\mathrm{pre}}$ and $\theta^*_{\mathrm{post}}$ to denote the parameters of the pre-trained and fine-tuned models.

\section{Method and Theoretical Results}
\label{sec:method:general}

\noindent \textbf{Problem setting.}
In our setting, we assume access to a pre-trained model $\theta^*_{\mathrm{pre}}$ but not necessarily to the pre-training dataset $\mathcal D_q$. Our goal is to adapt this model to capture as best as possible the distribution $p$, from which we only have access to a small dataset $\mathcal{D}_p = \{ \vx^{(k)}\}_{k=1}^{n_1}$. 

\subsection{\methodname \ algorithm}
\label{sec:method}
We start by presenting our algorithm right away.
\methodname{} fine-tunes a pre-trained model with parameters $\theta_{\mathrm{pre}}^*$ on the new dataset $\mathcal D_p$ with two losses on the same tokens: the usual cross-entropy on the observed next token, and a {\em temperature matching} loss that pulls the student's conditional distribution at temperature $T$ toward the pre-trained model's conditional distribution at temperature $T$. With $\mathcal D_p=\{\vx^{(k)}\}_{k=1}^{n_1}$ the post-training documents, each of length $l$, and $p_{T\theta}$ the temperature-scaled conditionals of Eq.~\eqref{eq:temp_scaled_conditionals}, the objective is
\begin{equation}
\widehat J_{\lambda,T}(\theta)=\frac{1}{n_1l}\sum_{k=1}^{n_1}\sum_{i=1}^{l}\Big[\underbrace{-\log p_\theta(x^{(k)}_i\mid\vx^{(k)}_{<i})}_{\text{Clean Cross-entropy}}+\lambda\underbrace{D_{\mathrm{KL}}\big(p_{T\theta^*_{\mathrm{pre}}}(\cdot\mid\vx^{(k)}_{<i})\,\|\,p_{T\theta}(\cdot\mid\vx^{(k)}_{<i})\big)}_{\text{Temperature Matching}}\Big]
\label{eq:tm-objective}
\end{equation}
where the student $\theta$ is initialized at $\theta^*_{\mathrm{pre}}$, the teacher $\theta^*_{\mathrm{pre}}$ is frozen, and $\lambda>0$ sets the relative weight of the two terms. The KL term equals the cross-entropy $-\sum_{y\in V}p_{T{\theta^*_{\mathrm{pre}}}}(y\mid\vx^{(k)}_{<i})\log p_{T\theta}(y\mid\vx^{(k)}_{<i})$ up to a constant that does not depend on $\theta$, which is what we optimize. 

We refer to the full objective of Eq.~\eqref{eq:tm-objective} as \methodname{} and to its second term as \tmname{}. The name connects to Ambient Diffusion~\citep{ambient_diffusion, consistent_diffusion_meets_tweedie, omni, dataloops}, a line of work that trains diffusion models from corrupted data. Here, we view as corruption the imperfect predictions of the pre-trained teacher, which we only trust at a high $T$.

\noindent \textbf{Extreme case 1: $T\to \infty$.} For any two distributions $p, q$ that have full support, the $D_{\mathrm{KL}}$ distance goes to $0$ as $T\to \infty$. Hence, the second term disappears, and our method reverts to regular SFT.

\noindent \textbf{Extreme case 2: $T=1$.} On the other hand,  setting $T=1$ gives rise to the classic Knowledge Distillation (KD) loss. Typically, in KD, the student is a weak model, and the teacher is a strong model. In our case, we have the opposite: the teacher is the pre-trained model, and the student is the model being post-trained that is expected to eventually outperform the teacher. 

\noindent \textbf{Practical Implementation.} Pseudocode is given in Algorithm~\ref{alg:tm}, Appendix~\ref{sec:appendix:algorithm}. Our implementation optimizes Eq. \eqref{eq:tm-objective} stochastically. At every step, with probability $p_c$ we use the regular SFT loss (clean cross-entropy only). With probability $1-pc$, we do a forward pass through the frozen teacher and we minimize the Temperature Matching loss between the tempered teacher and tempered student conditionals at every supervised position. In expectation, this implements \eqref{eq:tm-objective} with $\lambda=(1-p_c)/p_c$. 

\noindent \textbf{Computational overhead.} Since \methodname\ requires an additional forward pass through the frozen teacher, it is more expensive than the regular SFT. Using the standard approximation that a backward pass costs roughly twice a forward pass, the expected per-step compute relative to plain fine-tuning is $1 + \frac{1-p_c}{3} \cdot \frac{C_{\text{teacher}}}{C_{\text{student}}}$,
where $C_{\text{teacher}}/C_{\text{student}}$ is the ratio of teacher to student forward-pass cost. For our Experiments, we typically set $p_c = 0.75$ and we use the same model for the student and the teacher. This makes training $\approx 1.1\times$ slower for the same amount of steps. Hence, \methodname{} is best suited to settings that are \emph{data-limited} rather than \emph{compute-limited}.

\noindent \textbf{Clarification and Related Work.} Sometimes in the literature the term KD is used to refer to matching any teacher distribution, independent of the smoothing temperature $T$. In fact,~\citet{hinton2015distilling} experiment with matching at different temperatures. The main difference is that we use the pre-trained model to \emph{supervise itself} during post-training. An algorithm relevant to ours is the one from~\citet{li2017learning}. There a model is adapted to a new task while 
avoiding catastrophic forgetting of old tasks through regularization toward its own temperature-softened predictions on the previous tasks using new task's inputs. In LwF the new task has its own output head, so the distillation and new-task losses interact only through shared features. In our work, the output space is shared. This changes the role of temperature that is set to a fixed choice in LwF. In Ambient SFT, the temperature controls a bias-variance trade-off (Theorem~\ref{thm:loglinear-bias-variance}), and it is set based on how much the student trusts the base model. More discussion about related work is presented in Section~\ref{sec:appendix:related-work}.

\subsection{Theoretical Analysis of the Algorithm}
We now analyze \methodname{} theoretically: we show that pure temperature matching is equivalent to learning from high-temperature samples, characterize how fast temperature contracts distributional differences, and establish the statistical benefits of \methodname{} under a simple log-linear model. Readers interested mainly in the empirical results can skip to Section~\ref{sec:experiments}.

\subsubsection{Equivalence to learning from high-temperature data}
\label{sec:learning_high_temp_alone}
We start by showing that training with the temperature matching loss of Eq.~\eqref{eq:tm-objective} alone is equivalent to 
learning from high-temperature data alone. The result holds for matching toward \emph{any} autoregressive model, so we write $\theta^*$ for the teacher; in the algorithm $\theta^*=\theta^*_{\mathrm{pre}}$. The proof is in Appendix~\ref{sec:appendix:high-temp-proof}.

\begin{theorem}[Learning from noisy data]
\label{th:learning_from_high_temperature}
Let $p_{\theta^*}$ be any AR model over $V^l$ with logits $g_{\theta^*}$ as in Eq.~\eqref{eq:temp_1_conditionals}, and let $p_{T\theta^*}(\cdot\mid\vx_{<i})$ be its temperature-scaled conditionals of Eq.~\eqref{eq:temp_scaled_conditionals}. For every position $i$, let $\mu_i$ be any distribution over prefixes $\vx_{<i}\in V^{i-1}$.
\begin{compactenum}
\item[(a)] \textbf{\tmname{} is equivalent to learning from high-temperature samples.} Let $R_T(\theta)=\sum_{i=1}^{l}\mathbb E_{\vx_{<i}\sim\mu_i}\big[D_{\mathrm{KL}}\big(p_{T\theta^*}(\cdot\mid\vx_{<i})\,\|\,p_{T\theta}(\cdot\mid\vx_{<i})\big)\big]$ be the (population-level) temperature-matching loss of Eq.~\eqref{eq:tm-objective} with teacher $\theta^*$, and let
\begin{gather}
    J_T(\theta)=\sum_{i=1}^{l}\mathbb E_{\vx_{<i}\sim\mu_i}\,\mathbb E_{x_i\sim p_{T\theta^*}(\cdot\mid\vx_{<i})}\left[-\log p_{T\theta}(x_i\mid\vx_{<i})\right]
    \label{eq:ambient_loss_general}
\end{gather}
be the cross-entropy on next tokens sampled at temperature $T$. Then $J_T$, $R_T$ have the same minimizers. Unlike $R_T$, $J_T$ needs only samples from the tempered model, not its logits.
\item[(b)] \textbf{Exact recovery at any finite $T$.} If $\theta$ minimizes $J_T$ at any $T>0$, then $p_{T'\theta}(\cdot\mid\vx_{<i})=p_{T'\theta^*}(\cdot\mid\vx_{<i})$ for every context $\vx_{<i}$ in the support of $\mu_i$ and every $T'>0$, i.e., matching the teacher at any temperature implies matching the teacher at all temperatures. Moreover, this implies that $g_\theta(\cdot\mid\vx_{<i})=g_{\theta^*}(\cdot\mid\vx_{<i})+c(\vx_{<i})$ for a context-dependent constant. The prefix distributions only decide on which contexts the teacher is matched.
\end{compactenum}
\end{theorem}

Prefixes sampled from the tempered teacher are a special case (Corollary~\ref{cor:tempered-prefixes}, Appendix~\ref{sec:appendix:high-temp-proof}); in Eq.~\eqref{eq:tm-objective}, the prefixes come from the fine-tuning data. 
The theorem shows that the distribution is identifiable from high-temperature data. This is perhaps counterintuitive: at high $T$, samples look gibberish. Yet they still carry a learning signal, and a model trained with Eq.~\eqref{eq:ambient_loss} recovers the distribution, producing reasonable samples as shown in Figure~\ref{fig:example_training_data}.

\subsubsection{Temperature scaling as a contraction mechanism}
\label{sec:proxy}
If we had lots of data from $p_{\theta_{\mathrm{post}}^*}$, we could run temperature matching towards it for any $T$ of our choice. Unfortunately, in our setting we assume we do not have many data from it. What happens if we run temperature matching to the weak teacher $p_{\theta_{\mathrm{pre}}^*}$ instead? The following result establishes an upper bound on the KL divergence between the two distributions, for any fixed context as the temperature $T$ increases.
We show that, for every context $\vx_{<i}$ and every $T>0$:
\begin{align}
D_{\mathrm{KL}}\!\left(p_{T\theta_{\mathrm{post}}^*}(\cdot\mid \vx_{<i})\,\middle\|\,p_{T\theta_{\mathrm{pre}}^*}(\cdot\mid \vx_{<i})\right)\le\frac{\Delta(\vx_{<i})^2}{8T^2}, \label{eq:cos2}
\end{align}
where $\Delta(\vx_{<i})$ is the range over $y\in V$ of the logit gap $g_{\theta_{\mathrm{post}}^*}(y\mid \vx_{<i})-g_{\theta_{\mathrm{pre}}^*}(y\mid \vx_{<i})$. Our bound is stated in more detail in Appendix~\ref{sec:appendix:contraction_proof}. The bound gives a rate of $O(T^{-2})$ for every context. 

\subsubsection{\methodname{} analysis under a log-linear model}
\label{sec:learning_tradeoffs}

We are now ready to put all the pieces together.  Theorem~\ref{th:learning_from_high_temperature} establishes the equivalence of \tmname{} to learning from corrupted data from the teacher and shows that for any finite $T$, the teacher distribution is the minimizer of this term alone. Training with \tmname{} alone would be problematic since we would not be able to move from the pre-training distribution. Our algorithm trains with both regular SFT and Temperature Matching, with a choice of temperature $T$ controlling the level at which the gradients from the weak teacher can be reliable, as quantified by Eq. \eqref{eq:cos2}. The questions left to answer are: \textit{(a)} what is the right choice of temperature in the temperature matching term, and \textit{(b)} for the optimal amount of temperature, do we get any statistical advantage compared to learning from regular SFT alone? In what follows, we answer those questions. To avoid arguing about how LLMs generalize to new contexts not seen in training, we adopt a simple log-linear next-token model with frozen features, and analyze \eqref{eq:tm-objective} on it exactly.

\noindent \textbf{A log-linear model.}
Let $\phi: V \to \mathbb R^d$ be a fixed token embedding and $\psi: V^{*} \to \mathbb R^m$ a fixed context representation (e.g., the pre-trained network's last hidden state). Assume both functions have bounded $\ell_2$ norms. The learnable parameter $\theta$ is a single matrix, which we will also denote $A_\theta\in\mathbb R^{d \times m}$, and 
\begin{equation}
p_\theta(y\mid \vx_{<i})=\frac{\exp\big(\phi(y)^\top A_\theta\psi(\vx_{<i})\big)}{\sum_{z\in V}\exp\big(\phi(z)^\top A_\theta\psi(\vx_{<i})\big)}
\label{eq:loglinear-model}
\end{equation}
We can think of such a model as an architectural choice for $g_{\theta}$, i.e., we write $g_\theta(y\mid\vx_{<i})=\phi(y)^\top A_\theta\psi(\vx_{<i})$. Write $\Lambda(\eta)=\log\sum_{z\in V}\exp(\phi(z)^\top\eta)$ for the log-normalizer of \eqref{eq:loglinear-model}, i.e.\ the logarithm of its denominator, and $\mu(\eta)=\nabla\Lambda(\eta)=\mathbb E_{p_\eta}[\phi(y)]$, $\Sigma(\eta)=\nabla^2\Lambda(\eta)=\operatorname{Cov}_{p_\eta}[\phi(y)]$ for the mean embedding and the covariance of the conditional with natural parameter $\eta=A_\theta\psi(\vx_{<i})$. Flattening the sum of \eqref{eq:tm-objective}, the $n_1$ documents supply $n=n_1l$ token observations $(\vx_{<t},x_t)$, $t=1,\ldots,n$, and we write $\psi_t=\psi(\vx_{<t})$. We will use $\theta$ and $A_{\theta}$ interchangeably. 

\noindent \textbf{The objective.}
Instantiating \eqref{eq:tm-objective} on this model, 
gives, up to additive constants that do not depend on $\theta$,
\begin{equation}
\widehat J_{\lambda,T}(\theta)
=\underbrace{\frac1n\sum_{t=1}^{n}\Big[\Lambda(A_\theta\psi_t)-\phi(x_t)^\top A_\theta\psi_t\Big]}_{\widehat L(\theta)\ \text{(clean cross-entropy)}}
\;+\;\lambda\,\underbrace{\frac1n\sum_{t=1}^{n}\Big[\Lambda\!\big(\tfrac{A_\theta\psi_t}{T}\big)-\mu\!\big(\tfrac{A_{\theta_{\mathrm{pre}}^*}\psi_t}{T}\big)^{\!\top}\tfrac{A_\theta\psi_t}{T}\Big]}_{\widehat R_T(\theta)\ \text{(temperature matching)}} .
\label{eq:loglinear-objective}
\end{equation}

Write $\widehat\theta_\lambda$ for the minimizer of \eqref{eq:loglinear-objective}. Our goal in this section is to bound its distance from $\theta_{\rm post}^*$ and study the benefit of including temperature matching term in the objective. 

\noindent {\bf Remark:} For simplicity, we assume below that, in the data $(\vx_{<t},x_t)$, $t=1,\ldots,n$, that we have available for training, the prefixes $\vx_{<t}$ are drawn i.i.d.~from $p_{\theta^*_{\rm post}}$ and the tokens $x_t$ are also drawn independently from each other from $p_{\theta^*_{\rm post}}(\cdot | \vx_{<t})$. We denote with $\mathcal D_x$ the distribution of prefixes drawn from $p_{\theta^*_{\rm post}}$. Appendix~\ref{sec:appendix:loglinear} discusses how to deal with dependencies of tokens in documents.

Three quantities govern our result, where all expectations below are with respect to prefixes $\vx_{<t} \sim \mathcal D_x$, $\psi=\psi(\vx_{<t})$, and $\otimes$ is the Kronecker product:
\[
F=\mathbb E\big[\Sigma(A_{\theta^*_{\mathrm{post}}}\psi)\otimes\psi\psi^\top\big],
\quad
H_T=\frac1{T^2}\,\mathbb E\big[\Sigma(\tfrac{A_{\theta^*_{\mathrm{post}}}\psi}{T})\otimes\psi\psi^\top\big],
\quad
B=A_{\theta^*_{\mathrm{post}}}-A_{\theta^*_{\mathrm{pre}}}.
\]
We assume throughout that the frozen features are non-degenerate: $G=\mathbb E[\psi\psi^\top]\succ0$, and the embedding differences $\phi(y)-\phi(z)$ span $\mathbb R^d$. The latter fails only if some $v\neq0$ has $\phi(y)^\top v$ constant in $y$, in which case $A_\theta$ would not be identifiable; under the two conditions $F$ and $H_T$ are positive definite, so the inverses below are literal and $\|A_{\widehat\theta}-A_{\theta^*_{\mathrm{post}}}\|_F$ is a well-defined risk.

$F$ is the Fisher information matrix controlling the curvature of the first term in the loss function. 
It measures how much the target data tell us about each direction of $A_{\theta_{\rm post}^*}$. 

\noindent $H_T$ is the curvature of the temperature matching penalty term. As $T$ grows the tempered conditional approaches uniform, so $\Sigma(A_\theta\psi/T)\to\Sigma_0:=\operatorname{Cov}_{\mathrm{unif}(V)}[\phi(y)]$ for every $\theta$ and $H_T\approx\Sigma_0\otimes G/T^2$, where $G=\mathbb E[\psi\psi^\top]$, which is the same context geometry as $F$, but with a token covariance that never degenerates. The teacher thus adds curvature precisely where the data supplies none.

The next theorem quantifies how these quantities affect the estimation error.
The expansion concerns the regime in which the teacher's error and the statistical error of maximum likelihood are comparable, both of order $n^{-1/2}$, which is where the tradeoff between them is nontrivial. We therefore state the result under $\|B\|=O(n^{-1/2})$. The proof is in Appendix~\ref{sec:appendix:loglinear}. 

\begin{theorem}[Risk expansion and generic improvement over $T=1$]
\label{thm:loglinear-bias-variance}
Suppose $\|B\|=O(n^{-1/2})$. Then
\begin{align}
\mathbb E\big\|A_{\widehat\theta_{\lambda,T}}-A_{\theta^*_{\mathrm{post}}}\big\|_F^2&=\rho(\lambda,T)+O(n^{-3/2}),\nonumber\\
\rho(\lambda,T)&=\underbrace{\lambda^2\|(F+\lambda H_T)^{-1}H_TB\|_F^2}_{\text{bias}^2}
+\underbrace{\tfrac1n\operatorname{tr}\!\big((F+\lambda H_T)^{-1}F(F+\lambda H_T)^{-1}\big)}_{\text{variance}},
\label{eq:loglinear-risk}
\end{align}
uniformly for $\lambda$ and $T$ in a fixed range bounded away from $0$ and $\infty$. Moreover, interior choices of $(\lambda,T)$ do at least as well as all three extremes: 
\begin{equation}
\inf_{\substack{0<\lambda<\infty\\ 0<T<\infty,\ T\neq1}}\rho(\lambda,T)\ \le\
\min\Big\{\underbrace{\operatorname{tr}(F^{-1})/n}_{\text{maximum likelihood}},\ \underbrace{\|B\|_F^2}_{\text{frozen teacher}},\ \underbrace{\textstyle\min_{\lambda}\rho(\lambda,1)}_{\text{knowledge distillation}}\Big\},
\label{eq:loglinear-interior}
\end{equation}
the three terms inside the minimum being $\rho(0,T)$, $\lim_{\lambda\to\infty}\rho(\lambda,T)$ and the best $T=1$ run. Under a mild non-degeneracy condition, and for every $B$ outside a closed Lebesgue-null cone (see Appendix~\ref{sec:appendix:loglinear}), the inequality is strict: some finite $(\lambda,T)$ with $T\neq1$ strictly beats all three. 
\end{theorem}

\section{Experiments}
\label{sec:experiments}
\begin{figure*}[t]
\centering
\tcbset{fig2panel/.style={enhanced, equal height group=fig2, boxrule=0.5pt, arc=1.2mm,
  left=3pt, right=3pt, top=1.5pt, bottom=1.5pt, toptitle=0.5pt, bottomtitle=0.5pt,
  fonttitle=\scriptsize\sffamily\bfseries, coltitle=black}}
\begin{subfigure}[b]{0.60\linewidth}
\centering
\begin{tcolorbox}[fig2panel, width=0.475\linewidth, nobeforeafter, box align=top,
  colback=figdata!5, colframe=figdata!70!black, colbacktitle=figdata!20,
  title={Example training data}]
\raggedright\scriptsize\color{black!75}
The continued weird candidate Narrator Value complexities sheerRanked noveltyFighting diseaseibles condition Supplementary Maps gimm lol \dots
\end{tcolorbox}\hfill
\raisebox{-4ex}{\color{black!55}$\boldsymbol{\rightarrow}$}\hfill
\begin{tcolorbox}[fig2panel, width=0.475\linewidth, nobeforeafter, box align=top,
  colback=figgen!5, colframe=figgen!70!black, colbacktitle=figgen!20,
  title={Example (unconditional) generations}]
\raggedright\scriptsize
The structure should be:\\[0.2em]
1. First -- the software.\\
2. Second -- the database, which is usually used to store the authority of the code \dots
\end{tcolorbox}
\caption{Training data (left) and generations (right) of a model that has never seen a proper English sentence.}
\label{fig:example_training_data}
\end{subfigure}\hfill
\begin{subfigure}[b]{0.37\linewidth}
\centering
\includegraphics[width=\linewidth]{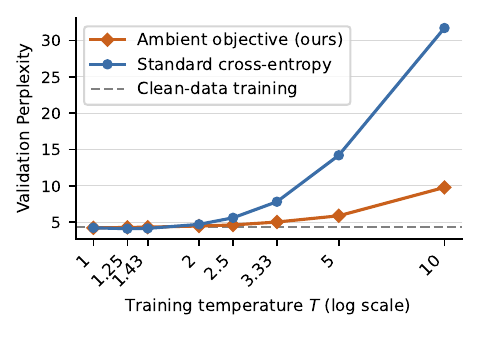}
\caption{Clean-data perplexity versus training temperature.}
\label{fig:shakespeare_dataset}
\end{subfigure}
\caption{Learning from high-temperature data.}
\label{fig:high-temperature-row}
\end{figure*}

\subsection{Learning from high-temperature data}
\label{sec:exp_noisy_oracle}

We first verify that training on high-temperature samples can still recover the original ($T=1$) distribution. We train a model on a dataset $\mathcal{D}$, sample from it at temperature $T$ to obtain $\mathcal{D}_T'$, train a new model on $\mathcal{D}_T'$ with Eq.~\eqref{eq:ambient_loss}, and measure its perplexity on a clean validation set.

For our first toy experiment, we use a GPT-2 architecture, and we train it on the tiny Shakespeare dataset~\citep{tinyshakespeare}. Figure~\ref{fig:shakespeare_dataset} shows the validation perplexity of a model trained with the objective of Eq.~\eqref{eq:ambient_loss} and a baseline trained with the normal loss function of Eq. \eqref{eq:normal_loss} for different temperatures. As shown, our algorithm achieves relatively low validation perplexity \emph{with respect to the clean distribution} even for high training corruption strengths. We proceed to repeat the idea at a larger scale. We train a GPT-2 124M model on OpenWebText ($\approx 9$B tokens), and we use it to produce synthetic data at different temperatures as before. Figure~\ref{fig:example_training_data} shows examples of training and generated data. The model is capable of producing clean samples even though it has only been trained with text that appears gibberish. 
These experiments validate that the high-temperature distributions still have usable learning signal, despite the seemingly uninformative training data.
Theorem \ref{th:learning_from_high_temperature} establishes that \tmname{} is exactly equivalent to learning from corrupted samples.

\subsection{Domain Adaptation Experiments}
\label{sec:domain-shift-experiments}

We proceed to post-training experiments, beginning with \emph{domain adaptation}: a model pre-trained on generic text needs to be adapted to a new domain from which we don't have much data.
We start with an $1.5$B-parameter model pre-trained on OpenWebText. We adapt to five target domains, chosen to span diverse text types: Python code~\citep{codeparrot_clean_train, chen2021evaluating}), CS papers~\citep{lo2020s2orc}, biomedical text~\citep{stuhlmann2025efficient}, news~\citep{Hamborg2017newspleaseA}, and product reviews~\citep{hou2026bridging}, each using $\approx$225M tokens. We benchmark: \textit{(a)} training from scratch on the domain of interest, \textit{(b)} fine-tuning, \textit{(c)} data-mixing (where we split training updates between the pre-training domain and the target domain), \textit{(d)} our \methodname{} approach. We always pick the hyperparameters that achieve the lowest validation perplexity in the target domain. Our algorithm has two hyperparameters $p_c$, which controls how often we apply the Temperature Matching loss and the temperature $T$. We give a full sweep of these parameters in Appendix \ref{sec:appendix:pt-grid}.

We report results in Table \ref{tab:domain-results-large}. As shown, \methodname{} matches or outperforms prior baselines across \textit{all} five domains. For News, the relative improvement in perplexity over the best baseline reaches $26.7\%$; in validation loss, this is a reduction from $1.99$ to $1.68$. 

\begin{table}[t]
\centering
\footnotesize
\caption{Validation losses achieved by different fine-tuning strategies for domain adaptation.}

\label{tab:domain-results-large}
\begin{tabular}{lccccc}
\toprule
Method & Python Code & CS & BioMed & News & Reviews \\
\midrule
Target Only (From Scratch) & 1.50 & 3.04 & 2.66 & 2.63 & 3.37 \\
Pre-trained (No Fine-tuning) & 1.46 & 2.87 & 2.61 & 2.59 & 3.56 \\
Target Only Fine-tuning     & 0.80 & 2.56 & 2.36 & 2.01 & \textbf{2.96} \\
Data Mixture Fine-tuning    & 0.79 & 2.55 & 2.36 & 1.99 & \textbf{2.96} \\
\hline
\methodname{} (ours)    & \textbf{0.70} & \textbf{2.54} & \textbf{2.35} & \textbf{1.68} & \textbf{2.96} \\
\bottomrule
\end{tabular}
\end{table}

\noindent \textbf{Predicting the success of \methodname.} The benefit varies across domains: it is largest on Python Code and News and smaller elsewhere. Because success depends on how much the teacher knows about the target domain, a cheap and reliable predictor of the benefit is the model's perplexity on the target domain before fine-tuning.
Across the five domains, this heuristic gives a Spearman rank correlation of $\rho=-0.90$ between zero-shot loss and win margin to the baselines. This is qualitatively consistent with Theorem~\ref{thm:loglinear-bias-variance}, whose bias term depends on the discrepancy $B=A_{\theta^*_{\mathrm{post}}}-A_{\theta^*_{\mathrm{pre}}}$. Full per-domain zero-shot loss and win-margin numbers are reported in Table~\ref{tab:zeroshot-domain-perplexity}.

\begin{figure}[t]
    \centering
    \begin{subfigure}[t]{0.49\linewidth}
        \centering
        \includegraphics[width=0.8\linewidth]{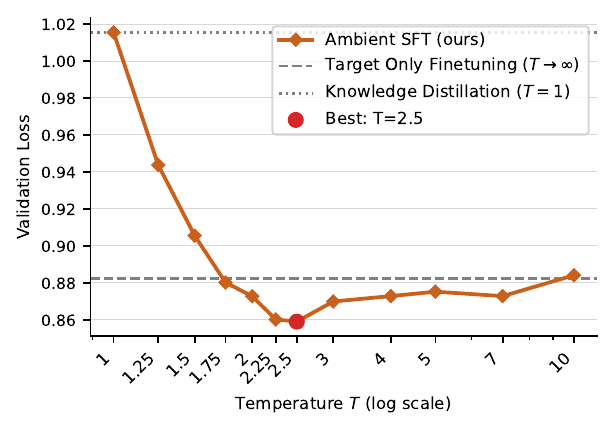}
        \caption{Effect of temperature $T$ on validation loss.}
        \label{fig:temperature_ablation}
    \end{subfigure}
    \hfill
    \begin{subfigure}[t]{0.49\linewidth}
        \centering
        \includegraphics[width=0.8\linewidth]{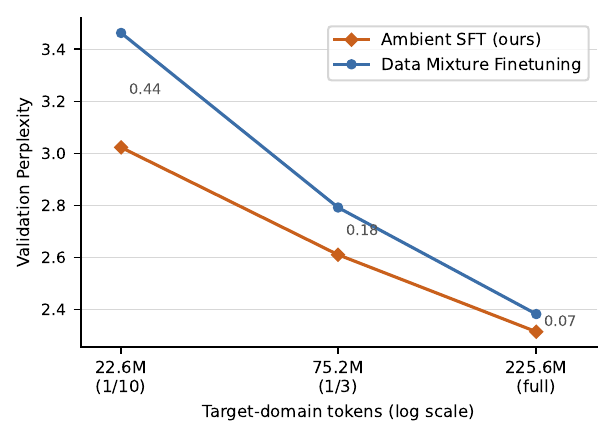}
        \caption{Effect of dataset size on validation perplexity.}
        \label{fig:data_size_ablation}
    \end{subfigure}
    \caption{Domain-adaptation ablations on Python Code.}
    \label{fig:domain-ablations}
\end{figure}

\noindent \textbf{Effect of temperature $T$.} The Temperature Matching term disappears as $T\to \infty$, and it becomes KD for $T=1$. Figure~\ref{fig:temperature_ablation} sweeps $T$, holding every other setting fixed. Neither extreme is optimal: validation loss falls steadily from $T=1$ down to a minimum at $T=2.5$, then rises and plateaus. Beyond this ablation, we clarify that we got the optimal $p_c$ and $T$ values used in Table \ref{tab:domain-results-large} by optimizing over the \emph{joint} parameter space. Hence, the optimal temperature $T^*$ satisfies $T^*\neq 1$.

\noindent \textbf{Effect of the size of the student and teacher.} In our experiments so far, the teacher is the frozen model after pre-training and the student is initialized from this teacher. Table~\ref{tab:teacher-strength-ablation} isolates the effect of student and teacher capacity, testing four combinations of student size (124M, 1.5B) and teacher size (124M, 1.5B). Two trends are unsurprising: \textit{(a)} a larger student improves results at both teacher sizes, and \textit{(b)} a larger teacher improves results at both student sizes. A third, less obvious finding is that a small (124M) teacher \emph{can} still improve a strictly larger (1.5B) student, a pairing that classical KD would not predict to help. Pairing the 1.5B student with the 124M teacher beats plain fine-tuning of the same 1.5B student with no teacher at all on Python Code and News. 

\noindent \textbf{Effect of number of tokens of the target domain.} To understand the effect of the size of the fine-tuning dataset, we subsample the Python Code training set, keeping only $1/3$ and $1/10$, and finetune a 124M model on each subsampled set with both \methodname{} and the Data Mixture baseline. Figure~\ref{fig:data_size_ablation} reports the results: our method consistently outperforms the baseline, and the gap widens as the dataset shrinks. This is intuitive: with enough target-domain data, our method should not help, but it is highly effective when data, not compute, is the bottleneck. This matches Theorem~\ref{thm:loglinear-bias-variance}: the bias term is independent of $n$, while the sampling-variance term is not.

\subsection{Instruction-tuning experiments}
\label{sec:instruct}
\label{sec:instruct-experiments}
Having established \methodname{} as a drop-in replacement for SFT in domain adaptation, we now ask whether it helps elsewhere in the post-training stack, starting with instruction tuning, where a pre-trained LLM is trained via SFT to follow instructions. We start with pre-trained base models, ranging from $1.7$B to $14$B params, and we fine-tune them on the Dolci-Instruct-SFT~\citep{olmo2025olmo}. 
We start our evaluation by using only $\approx 20$K documents  ($\approx 18$M tokens). 
We validate performance on standard benchmarks that measure instruction following (IFEval~\citep{zhou2023instruction}), knowledge (MMLU~\citep{hendrycks2020measuring}, PopQA~\citep{mallen2023not}, TriviaQA~\citep{joshi2017triviaqa}), math (GSM8K~\citep{cobbe2021training}, MATH~\citep{hendrycks2021measuring}), code (HumanEval+~\citep{chen2021evaluating}), and general reasoning (BBH~\citep{suzgun2023challenging}). For all our runs, we use $p_c=0.75$ (i.e., $25\%$ of the updates are spent on the \tmname{} term) and $T=2.5$. 

Results are presented in Table \ref{tab:instruct-lowdata}, where we show results for Qwen3 models ranging from 1.7B to 14B parameters. We start by analyzing what Vanilla SFT does to the capabilities of the underlying model. As shown, Vanilla SFT improves instruction-following capabilities by a lot (e.g. IFEval increases from $45.5\%$ to $76.3\%$ at the 4B scale) but leads to catastrophic forgetting on the rest of the benchmarks: the base model (pre-SFT) beats vanilla-SFT on 6 or 7 tasks for every model. On the contrary,
\methodname{} leads to consistent average improvements ranging from $8.0\%$ to $20.3\%$. At every size and model family, \methodname{} matches or beats vanilla SFT on every benchmark except IFEval, and significantly outperforms the base model on IFEval. 
Beyond that, \methodname{} mostly \textbf{preserves knowledge} (MMLU, PopQA, TriviaQA), and \textbf{increases capabilities in math}: it is above the base model on both GSM8K and MATH for almost every model tested.

\begin{table}[t]
\centering
\caption{Instruction tuning on $\approx21$K examples, $1\%$ of Dolci-Instruct-SFT, \small{Qwen3 base models from 1.7B to 14B parameters. \methodname{} improves on Vanilla SFT on 7/8 tasks, consistently across sizes; the exception is IFEval, where the gap closes as the model grows. Last column is the average relative improvement.
}}
\label{tab:instruct-lowdata}
\footnotesize
\setlength{\tabcolsep}{3.5pt}
\resizebox{\linewidth}{!}{%
\begin{tabular}{llcccccccc|c}
\toprule
 & & MATH & GSM8K & \makecell{HumanEval+} & MMLU & PopQA & TriviaQA & IFEval & BBH & \makecell{Avg.\ rel.\ $\Delta$\\vs.\ SFT} \\
\midrule
\multirow{3}{*}{Qwen3-1.7B-Base} & Base (no SFT) & 33.8 & 53.5 & 78.4 & \textbf{65.6} & 13.9 & \textbf{45.6} & 37.9 & \textbf{54.2} & $+9.0\%$ \\
 & Vanilla SFT & 24.3 & 61.0 & 68.0 & 61.9 & 12.2 & 36.3 & \textbf{58.8} & 45.2 & --- \\
 & \methodname{} & \textbf{40.4} & \textbf{72.3} & \textbf{83.7} & 64.9 & \textbf{14.9} & 43.8 & 52.3 & 53.3 & $+20.3\%$ \\
\midrule
\multirow{3}{*}{Qwen3-4B-Base} & Base (no SFT) & \textbf{43.5} & 81.9 & \textbf{91.2} & \textbf{75.6} & \textbf{18.4} & \textbf{60.4} & 45.5 & \textbf{70.5} & $+5.7\%$ \\
 & Vanilla SFT & 36.4 & 80.2 & 84.5 & 72.9 & 14.8 & 51.4 & \textbf{76.3} & 63.4 & --- \\
 & \methodname{} & 42.2 & \textbf{82.9} & 90.1 & 74.3 & 18.1 & 57.5 & 71.0 & 68.9 & $+8.0\%$ \\
\midrule
\multirow{3}{*}{Qwen3-8B-Base} & Base (no SFT) & 48.5 & 85.8 & \textbf{93.1} & \textbf{78.8} & \textbf{24.8} & \textbf{70.7} & 56.2 & \textbf{76.1} & $+9.1\%$ \\
 & Vanilla SFT & 38.7 & 80.4 & 82.5 & 76.2 & 18.2 & 65.2 & \textbf{77.1} & 71.4 & --- \\
 & \methodname{} & \textbf{53.5} & \textbf{87.5} & 91.5 & 78.3 & 22.0 & 69.2 & 71.7 & 72.0 & $+10.2\%$ \\
\midrule
\multirow{3}{*}{Qwen3-14B-Base} & Base (no SFT) & 51.2 & 81.5 & \textbf{93.7} & \textbf{82.2} & \textbf{28.9} & \textbf{77.0} & 68.0 & 76.9 & $+6.3\%$ \\
 & Vanilla SFT & 42.8 & 86.7 & 86.9 & 80.9 & 21.8 & 71.6 & \textbf{78.6} & 76.2 & --- \\
 & \methodname{} & \textbf{58.9} & \textbf{89.3} & 92.6 & 81.1 & 24.7 & 75.5 & \textbf{78.6} & \textbf{78.5} & $+8.7\%$ \\
\bottomrule
\end{tabular}}
\end{table}

\begin{table}[!htp]
\centering
\caption{\methodname{} with a single temperature ($T=2.5$) and with a temperature set per source.}
\label{tab:per-source-T}
\footnotesize
\setlength{\tabcolsep}{3.5pt}
\resizebox{\linewidth}{!}{%
\begin{tabular}{llcccccccc|c}
\toprule
 & & MATH & GSM8K & \makecell{HumanEval+} & MMLU & PopQA & TriviaQA & IFEval & BBH & \makecell{Avg.\ rel.\ $\Delta$\\vs.\ SFT} \\
\midrule
\multirow{3}{*}{Qwen3-1.7B} & Vanilla SFT & 24.3 & 61.0 & 68.0 & 61.9 & 12.2 & 36.3 & \textbf{58.8} & 45.2 & --- \\
 & \methodname{} & 40.4 & 72.3 & \textbf{83.7} & \textbf{64.9} & \textbf{14.9} & \textbf{43.8} & 52.3 & \textbf{53.3} & $+20.3\%$ \\
 & \methodname{} (per-source $T$) & \textbf{40.9} & \textbf{73.2} & 83.1 & 64.5 & 14.6 & 42.8 & 58.2 & 53.2 & $+21.1\%$ \\
\midrule
\multirow{3}{*}{Qwen3-4B} & Vanilla SFT & 36.4 & 80.2 & 84.5 & 72.9 & 14.8 & 51.4 & 76.3 & 63.4 & --- \\
 & \methodname{} & 42.2 & 82.9 & 90.1 & \textbf{74.3} & \textbf{18.1} & \textbf{57.5} & 71.0 & 68.9 & $+8.0\%$ \\
 & \methodname{} (per-source $T$) & \textbf{43.4} & \textbf{85.3} & \textbf{90.6} & 73.7 & 16.7 & 55.6 & \textbf{77.1} & \textbf{69.8} & $+8.2\%$ \\
\midrule
\multirow{3}{*}{Qwen3-8B} & Vanilla SFT & 38.7 & 80.4 & 82.5 & 76.2 & 18.2 & 65.2 & 77.1 & 71.4 & --- \\
 & \methodname{} & 53.5 & \textbf{87.5} & 91.5 & \textbf{78.3} & \textbf{22.0} & \textbf{69.2} & 71.7 & \textbf{72.0} & $+10.2\%$ \\
 & \methodname{} (per-source $T$) & \textbf{54.5} & 85.2 & \textbf{92.7} & 77.2 & 20.7 & 68.1 & \textbf{78.6} & 71.5 & $+10.1\%$ \\
\bottomrule
\end{tabular}}
\end{table}

\begin{table}[!htp]
\centering
\caption{Generalization across model families, in the same setup as Table~\ref{tab:instruct-lowdata}.}
\label{tab:instruct-families}
\footnotesize
\setlength{\tabcolsep}{3.5pt}
\resizebox{\linewidth}{!}{%
\begin{tabular}{llcccccccc|c}
\toprule
 & & MATH & GSM8K & \makecell{HumanEval+} & MMLU & PopQA & TriviaQA & IFEval & BBH & \makecell{Avg.\ rel.\ $\Delta$\\vs.\ SFT} \\
\midrule
\multirow{2}{*}{OLMo-3-7B} & Vanilla SFT & 22.1 & 75.2 & 71.4 & 65.7 & 17.5 & 60.9 & \textbf{70.8} & 61.3 & --- \\
 & \methodname{} & \textbf{41.5} & \textbf{78.6} & \textbf{79.5} & \textbf{66.4} & \textbf{21.0} & \textbf{64.8} & 64.0 & \textbf{62.4} & $+15.4\%$ \\
\midrule
\multirow{2}{*}{Llama-3.1-8B} & Vanilla SFT & 17.8 & 58.8 & 55.2 & 63.2 & 26.9 & 75.1 & \textbf{71.9} & 61.4 & --- \\
 & \methodname{} & \textbf{21.2} & \textbf{61.1} & \textbf{70.5} & \textbf{65.7} & \textbf{30.9} & \textbf{79.3} & 62.7 & \textbf{64.1} & $+8.3\%$ \\
\bottomrule
\end{tabular}}
\end{table}
\noindent \textbf{Effect of model size.} The gain is largest at the smallest scale but persists as models grow: the average relative improvement is $+20.3\%$ at 1.7B and stays above $+8.0\%$ from 4B to 14B. On the eight-benchmark average, \methodname{} at 8B equals vanilla SFT at 14B, and on MATH \methodname{} at 1.7B already exceeds vanilla SFT at 8B. At 14B, Ambient SFT ties Vanilla SFT on IFEval.

\noindent \textbf{Per-source temperature.} 
The only benchmark where Ambient SFT is worse than vanilla SFT is IFEval. We find the problem originates from the composition of our dataset. Dolci-Instruct-SFT mixes 21 sources, and the base model fits some of them much better than others.
Setting the temperature separately for each source improves IFEval at every model size, while keeping other gains. Results are shown in Table~\ref{tab:per-source-T} and details on how to set the temperature for each source are given in Appendix~\ref{sec:appendix:per-source-T}. 

\noindent \textbf{Generalization to other model families.} Table \ref{tab:instruct-families} shows more model families for which Ambient SFT outperforms vanilla SFT. Specifically, we show large gains for OLMo-3-7B and Llama-3.1-8B.

\begin{table}[t]
\centering
\caption{Comparison with other regularizers, Qwen3-1.7B-Base, Table~\ref{tab:instruct-lowdata} setup. \small{Full sweeps in Table~\ref{tab:baselines-full}.}}
\label{tab:baselines}
\footnotesize
\setlength{\tabcolsep}{3.5pt}
\resizebox{\linewidth}{!}{%
\begin{tabular}{lcccccccc|c}
\toprule
& MATH & GSM8K & \makecell{HumanEval+} & MMLU & PopQA & TriviaQA & IFEval & BBH & \makecell{Avg.\ rel.\ $\Delta$\\vs.\ SFT} \\
\midrule
Base (no SFT) & 33.8 & 53.5 & 78.4 & \textbf{65.6} & 13.9 & \textbf{45.6} & 37.9 & \textbf{54.2} & $+9.0\%$ \\
\midrule
Vanilla SFT & 24.3 & 61.0 & 68.0 & 61.9 & 12.2 & 36.3 & 58.8 & 45.2 & --- \\
NEFTune ($\alpha=10$) & 19.7 & 58.9 & 69.2 & 62.2 & 12.4 & 38.4 & \textbf{65.6} & 48.6 & $+0.8\%$ \\
L2-SP ($\alpha=1$) & 35.0 & \textbf{72.3} & 83.2 & 64.9 & 14.5 & \textbf{45.6} & 41.8 & 52.8 & $+15.3\%$ \\
Label smoothing ($\epsilon=0.2$) & 25.4 & 62.9 & 64.3 & 60.9 & 12.1 & 36.4 & 55.1 & 47.0 & $-0.3\%$ \\
KD ($T=1$, $p_c=0.5$) & 35.1 & 71.5 & 81.3 & 64.8 & \textbf{15.7} & 43.8 & 49.2 & 52.8 & $+17.0\%$ \\
\midrule
\methodname{} & \textbf{40.4} & \textbf{72.3} & \textbf{83.7} & 64.9 & 14.9 & 43.8 & 52.3 & 53.3 & $+20.3\%$ \\
\bottomrule
\end{tabular}}
\end{table}

\noindent \textbf{Effect of dataset size.} We now study what happens when we use $1\%, 10\%$ and $100\%$ of the dataset. The greatest improvements are observed in the $1\%$ of the dataset setting, as expected, with an average $15.4\%$ relative improvement over vanilla SFT; with $10\%$ of the dataset, the average relative improvement is $3.4\%$. That said, \textit{even at full data scale}, \methodname{} leads to an average $8.6\%$ relative improvement to vanilla SFT. Figure~\ref{fig:data-scaling} shows per-benchmark scores across different scales.

\begin{figure}[!htp]
\centering
\includegraphics[width=0.8\linewidth]{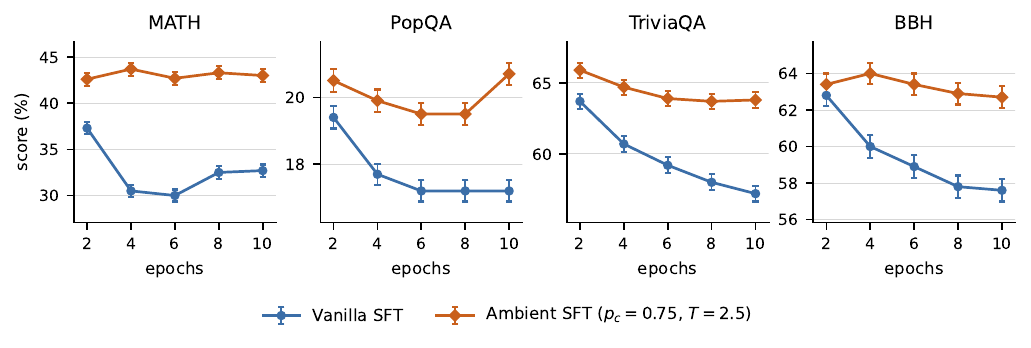}
\caption{Results versus $\#$ of epochs following instruction-tuning on $10\%$ of Dolci-Instruct-SFT.}
\label{fig:epoch-sweep}
\end{figure}

\noindent \textbf{\methodname{} reduces overfitting.} We hypothesize that \methodname{} reduces overfitting, letting the model benefit from more epochs than vanilla SFT. To test this, we train for $2$, $4$, $6$, $8$, and $10$ epochs on a $10\%$ subset of Dolci-Instruct-SFT (215k examples). As Figure~\ref{fig:epoch-sweep} shows, \methodname{} reaches its best performance at more epochs and is more robust to overfitting.

\noindent \textbf{Vanilla SFT trades knowledge for instruction following.} For vanilla SFT, the number of epochs controls a trade-off. Going from $2$ to $6$ epochs raises IFEval from $71.0$ to $80.8$, but lowers every other benchmark: MATH from $37.3$ to $30.0$, GSM8K from $76.4$ to $69.6$, HumanEval+ from $81.3$ to $75.5$, MMLU from $66.3$ to $65.0$, PopQA from $19.4$ to $17.2$, TriviaQA from $63.7$ to $59.2$ and BBH from $62.8$ to $58.9$. Stopping early therefore limits forgetting only by giving up instruction following. \methodname{} does not face this trade-off: over the same range, IFEval rises from $63.8$ to $73.4$, while the other benchmarks stay within about two points of their 2-epoch values.

\noindent \textbf{Comparison with other regularizers.} Our Temperature Matching term can be thought as regularization. To compare with other regularizers, Table~\ref{tab:baselines} shows results on Qwen3-1.7B with L2-SP~\citep{xuhong2018explicit}, NEFTune~\citep{jain2310neftune}, KD toward the frozen base model at $T=1$, and label smoothing, each at its best setting. \methodname{} outperforms other regularization methods. It reaches $+20.3\%$ over vanilla SFT, with the second best method being KD (Ambient SFT with $T=1$) with $+17.0\%$.

\subsection{Reasoning experiments}
Finally, we test whether \methodname{} benefits reasoning. We train OLMo-3-7B on $\approx 22$K examples ($1\%$) of Dolci-Think-SFT~\citep{olmo2025olmo} with the same hyperparameters as before. We evaluate on GSM8K~\citep{cobbe2021training}, AIME 2024 and 2025, HumanEval+~\citep{chen2021evaluating}, IFEval~\citep{zhou2023instruction}, and GPQA~\citep{rein2023gpqa} (details in Appendix~\ref{sec:appendix:harness-reasoning}). As Table~\ref{tab:reasoning-baselines} shows, \methodname{} outperforms vanilla SFT on 5/6 benchmarks and trails by only $1.2$ points on GSM8K. While there is a consistent boost in reasoning abilities, its magnitude is smaller compared to the one observed for domain adaptation and instruction-tuning. That said, Table~\ref{tab:reasoning-knowledge} further shows that reasoning models trained with Ambient SFT are also more robust to prompting than pure SFT ones. Ambient SFT also is the only regularization method that leads to improvements over SFT for reasoning, as both KD and L2-SP lead to regression in the reasoning scores.

\begin{table}[!htb]
\centering
\caption{Reasoning (OLMo-3-7B, $1\%$ of Dolci-Think-SFT): \methodname{} against Vanilla SFT and the two strongest baselines of Table~\ref{tab:baselines}, at their best settings.}
\label{tab:reasoning-baselines}
\footnotesize
\resizebox{\linewidth}{!}{%
\begin{tabular}{lcccccc|c}
\toprule
 & GSM8K & AIME24 & AIME25 & GPQA & IFEval & HumanEval+ & \makecell{Avg.\ rel.\ $\Delta$\\vs.\ SFT} \\
\midrule
Base (no SFT) & 85.5 & 10.6 & 13.9 & 27.5 & 33.5 & 45.9 & $-35.6\%$ \\
Vanilla SFT & \textbf{94.2} & 39.8 & 33.2 & 39.3 & 34.6 & 75.6 & --- \\
KD ($T=1$, $p_c=0.5$) & 90.7 & 36.2 & 30.8 & 33.0 & 34.0 & 70.7 & $-7.3\%$ \\
L2-SP ($\alpha=1$) & 81.7 & 9.9 & 14.0 & 28.6 & 33.8 & 40.3 & $-37.1\%$ \\
\midrule
\methodname{} & 93.0 & \textbf{44.5} & \textbf{34.1} & \textbf{40.0} & \textbf{36.0} & \textbf{76.9} & $+3.5\%$ \\
\bottomrule
\end{tabular}}
\end{table}

\section{Related Work}
\label{sec:appendix:related-work}
Our work builds upon the Knowledge Distillation (KD) literature pioneered by~\citet{bucilua006model} and~\citet{hinton2015distilling}. In the context of LLMs, KD has been successfully applied in the works of~\citep{kim2016sequence, gu2023minillm, zhang2019your}. However, whereas classical KD employs a teacher that is stronger or more aligned than the student, our framework inverts this dynamic: the teacher model is strictly \textit{weaker} at the target task. Because the student (the fine-tuned model) is expected to outperform the teacher (the frozen pre-trained model), the teacher's role is not to act as an oracle. Instead, it serves to regularize the fine-tuning process by providing coarse, high-temperature supervision. We also note that applying KD on smoothed distributions has been explored in the paper by~\citep{hinton2015distilling} and an explicit scaling term $1/T^2$ has been proposed. In our experiments, we do not find this scaling to be optimal across all of our settings, and we apply Knowledge Distillation to a previous version of the same model instead of an external more powerful teacher. The concept of self-distillation has also been explored in~\citep{yang2024self}. The main difference between this work and ours is that we match towards the full high-temperature distribution of the teacher, while the authors of~\citep{yang2024self} run SFT over datasets produced by the model.

Closer to our setting is the continual-learning literature on avoiding catastrophic forgetting without access to the original training data.~\citet{li2017learning} propose recording the pre-update network's outputs on the new task's own inputs and adds a temperature-softened distillation term to the new task's loss, anchoring the update to the model's previous behavior without replaying old data. Our method shares that structure: the frozen pre-trained model is queried on the fine-tuning data itself, and the match is enforced after raising the temperature.

The concept of anchoring a model to a prior state also frequently appears in reinforcement learning (RL) alignment pipelines, where a KL-divergence penalty against a reference model is used to prevent reward hacking~\citep{ouyang2022training}. Our framework diverges from this paradigm in two critical ways. First, our method targets the Supervised Fine-Tuning (SFT) stage rather than the RL phase. Second, we exclusively match the high-temperature distributions of the reference model. Enforcing a strict match at $T=1$ would actively degrade performance by forcing the student to adopt the unaligned base model's lack of task awareness, a manifestation of the phenomenon known as alignment tax~\citep{ouyang2022training}. By matching the reference model only at a coarse level, our method successfully avoids this tax while maintaining structural regularization. Table~\ref{tab:alignment_taxonomy} in Appendix~\ref{sec:appendix:prior-methods} summarizes how \methodname{} differs from these techniques.

\section{Conclusion and Future Work}
We introduced \methodname{}, a drop-in replacement for SFT that adds to the standard cross-entropy loss a term matching the model to a high-temperature version of its own pre-trained predictions. Across domain adaptation, instruction tuning, and reasoning, \methodname{} outperforms vanilla SFT, with the largest gains when fine-tuning data are scarce. A natural next step is to extend this anchoring idea beyond SFT to the reinforcement-learning algorithms used in post-training.

\FloatBarrier
\subsection*{Acknowledgments}
Giannis Daras is supported by a Jane Street Research Collaboration award, a TPU Research Cloud grant, compute grants from Lambda Research and Thinking Machines, and API credits from the Anthropic AI for Science program. Antonio Torralba is supported by ONR MURI grant \mbox{N00014-26-1-2255}. Constantinos Daskalakis is supported by a Simons Investigator Award, the Simons Collaboration on the Theory of Algorithmic Fairness, ONR MURI grant \mbox{N00014-25-1-2116}, and ONR grant \mbox{N00014-25-1-2296}. The experiments in this work were run on the Vista supercomputer at the Texas Advanced Computing Center (TACC), on the Engaging cluster of the MIT Office of Research Computing and Data (ORCD), and on Google Cloud TPUs provided by the TPU Research Cloud.

\subsection*{AI Use Statement}
In this work, we used generative AI tools for assistance with paper writing, editing figures, developing the mathematical theorems and proofs, implementing parts of the method and conducting the experiments.
We have \textbf{not} used generative AI tools for conceptualization of the project, results interpretation and developing the structure and the narrative of the paper. We have reviewed all AI-assisted work. In particular, we reviewed the code, the proofs, all the paper edits and parsing of the results of the experiments. The authors fully understand and take ownership of all the materials presented in this work. We take responsibility for the final content of this work, including text, claims or
artifacts produced with the aid of generative AI.

\bibliographystyle{iclr2027_conference}
\bibliography{references,ambient_and_friends}

\appendix
\clearpage
\addtocontents{toc}{\protect\setcounter{tocdepth}{3}}
\renewcommand{\contentsname}{Appendix: contents}
{\hypersetup{hidelinks}\tableofcontents}
\clearpage

\section{Comparison with Prior Techniques}
\label{sec:appendix:prior-methods}
\begin{table}[H]
\centering
\footnotesize
\caption{Comparison of our technique (\methodname) with existing techniques in the literature. Our method targets the SFT part of the post-training process, and it does \textbf{not} rely on model rollouts. It is using a noisy teacher (model before fine-tuning) during fine-tuning. To account for the imperfections of the teacher, we only use it at high-temperatures.}
\label{tab:alignment_taxonomy}
\begin{tabular}{@{}>{\raggedright\arraybackslash}p{0.235\linewidth} >{\raggedright\arraybackslash}p{0.205\linewidth} >{\raggedright\arraybackslash}p{0.265\linewidth} >{\raggedright\arraybackslash}p{0.205\linewidth}@{}}
\toprule
\textbf{Method} & \textbf{Input Context ($x$) Source} & \textbf{Target ($y$) Source} & \textbf{Loss Type} \\
\midrule
\rowcolor{sftgreen} \textbf{Standard SFT} & Fine-tuning Dataset & Dataset / Human & Hard labels \\
\rowcolor{sftgreen} \textbf{SFT Teacher Data} & Teacher Rollouts & Teacher Final Output & Hard labels \\
\rowcolor{sftgreen} \textbf{Knowledge Distillation} & Fine-tuning Dataset & Teacher Logits & Soft labels \\
\rowcolor{rlblue} \textbf{On-Policy Distillation} & Student Rollouts & Teacher Rewards & RL \\
\rowcolor{rlblue} \textbf{Off-Policy Distillation} & Teacher Rollouts & Teacher Rewards & RL \\
\midrule
\rowcolor{sftgreen} \textbf{\methodname} (Ours) & Fine-tuning Dataset & \textbf{Pre-Trained Model (Same Size)} & \textbf{Soft at high-temperature} \\
\bottomrule
\end{tabular}
\end{table}

\section{Training Algorithm}
\label{sec:appendix:algorithm}

Algorithm~\ref{alg:tm} is the stochastic implementation of the objective \eqref{eq:tm-objective} of Section~\ref{sec:method}. Each step draws one rank-synchronized coin: with probability $p_c$ the batch takes a standard SFT step, and otherwise the frozen teacher is run forward without gradients and the step minimizes the matching loss at every supervised position. In expectation this is \eqref{eq:tm-objective} with $\lambda=(1-p_c)/p_c$, and $p_c=1$ recovers SFT exactly.

\begin{algorithm}[t]
\caption{\methodname{} (drop-in replacement for SFT)}
\small
\label{alg:tm}
\begin{algorithmic}[1]
\Require pre-trained weights $\theta_{\mathrm{pre}}^*$; post-training documents $\mathcal D_p$; clean probability $p_c\in(0,1]$; temperature $T>0$; number of steps $S$; optimizer \textsc{Opt}
\State $\theta\gets\theta^*_{\mathrm{pre}}$ (student); keep a frozen copy of $\theta^*_{\mathrm{pre}}$ (teacher)
\For{$s=1,\dots,S$}
  \State sample a batch $\mathcal B\subset\mathcal D_p$; let $\mathcal I$ be its supervised positions $(k,i)$ and $N=|\mathcal I|$
  \State $g_{(k,i)}\gets g_\theta(\cdot\mid\vx^{(k)}_{<i})$ for $(k,i)\in\mathcal I$ \Comment{student logits, one forward pass}
  \State draw $c\sim\mathrm{Bernoulli}(p_c)$ \Comment{shared across data-parallel ranks}
  \If{$c=1$} \Comment{clean SFT step}
    \State $\mathcal L\gets-\frac1N\sum_{(k,i)\in\mathcal I}\log\mathrm{softmax}(g_{(k,i)}({x^{(k)}_i}))$
  \Else \Comment{temperature-matching step}
    \State $\tilde g_{(k,i)}\gets g_{\theta^*_{\mathrm{pre}}}(\cdot\mid\vx^{(k)}_{<i})$ for $(k,i)\in\mathcal I$, without gradients \Comment{teacher logits}
    \State $q_{(k,i)}\gets\mathrm{softmax}(\tilde g_{(k,i)}/T)$ \Comment{tempered teacher}
    \State $\mathcal L\gets-\frac1N\sum_{(k,i)\in\mathcal I}\sum_{y\in V}q_{(k,i)}(y)\,\log\mathrm{softmax}(g_{(k,i)}(y)/T)$ \Comment{CE to tempered teacher}
  \EndIf
  \State $\theta\gets\textsc{Opt}(\theta,\nabla_\theta\mathcal L)$
\EndFor
\State \Return $\theta$
\end{algorithmic}
\end{algorithm}

\section{Experimental Details}
\label{sec:appendix:experimental-details}

\subsection{Domain Adaptation}
\label{sec:appendix:details-domain}

\subsubsection{Datasets}
\label{sec:appendix:datasets}

All models are pre-trained on OpenWebText ($\approx$9B tokens), a publicly available reproduction of the WebText corpus used to train GPT-2, obtained from the \texttt{Skylion007/openwebtext}\footnote{\url{https://huggingface.co/datasets/Skylion007/openwebtext}} Hugging Face dataset. For the domain adaptation experiments of Section~\ref{sec:domain-shift-experiments}, we adapt these pre-trained models to five target domains, chosen to span a diverse range of text types. Each domain corpus is capped at $\approx$225M target-domain tokens, matching the same order of magnitude across domains so that no single domain benefits from a systematically larger fine-tuning set. The domains, their sources, and the exact Hugging Face datasets used are:

\begin{itemize}
    \item \textbf{Python Code}: an $\approx 225$M token subset of the CodeParrot clean-train dataset~\citep{codeparrot_clean_train}, itself built from permissively-licensed GitHub Python repositories in the style of the training corpus used by~\citet{chen2021evaluating}. \texttt{codeparrot/codeparrot-clean-train}\footnote{\url{https://huggingface.co/datasets/codeparrot/codeparrot-clean-train}}.
    \item \textbf{CS}: computer science paper abstracts drawn from S2ORC, the Semantic Scholar Open Research Corpus~\citep{lo2020s2orc}. \texttt{AlgorithmicResearchGroup/s2orc-cs-enriched}\footnote{\url{https://huggingface.co/datasets/AlgorithmicResearchGroup/s2orc-cs-enriched}}.
    \item \textbf{BioMed}: biomedical abstracts drawn from PubMed metadata (National Library of Medicine), via the subset released alongside~\citet{stuhlmann2025efficient}. \texttt{slinusc/PubMedAbstractsSubset}\footnote{\url{https://huggingface.co/datasets/slinusc/PubMedAbstractsSubset}}.
    \item \textbf{News}: English-language news articles from CC-News, a Common Crawl news dataset extracted with the \texttt{news-please} crawler~\citep{Hamborg2017newspleaseA}. \texttt{vblagoje/cc\_news}\footnote{\url{https://huggingface.co/datasets/vblagoje/cc_news}}.
    \item \textbf{Reviews}: Amazon product reviews (books category), drawn from the Amazon Reviews 2023 dataset~\citep{hou2026bridging}. \texttt{cogsci13/\allowbreak Amazon-\allowbreak Reviews-\allowbreak 2023-\allowbreak Books-\allowbreak Review}\footnote{\href{https://huggingface.co/datasets/cogsci13/Amazon-Reviews-2023-Books-Review}{\texttt{huggingface.co/\allowbreak datasets/\allowbreak cogsci13/\allowbreak Amazon-\allowbreak Reviews-\allowbreak 2023-\allowbreak Books-\allowbreak Review}}}.
\end{itemize}

\subsubsection{Full $(p,T)$ sweep and optimal hyperparameters}
\label{sec:appendix:pt-grid}

Table~\ref{tab:optimal-pt-small-student} reports the optimal $(p,T)$ per domain used to produce Table~\ref{tab:domain-results-small} and Table~\ref{tab:teacher-strength-ablation}'s 124M-student rows. The optimal mixing ratio $p$ was $0.5$ in every one of the 10 cells tested, so the table reports only $T$.

For the Small Teacher condition specifically, Tables~\ref{tab:pt-grid-cs}--\ref{tab:pt-grid-pythoncode} report the full $(p,T)$ grid each domain's optimum was selected from: a uniform 15-point grid ($p\in\{0.5,0.75,0.9\}\times T\in\{1.25,1.5,1.75,2.0,2.5\}$) for CS, BioMed, News, and Reviews, and a wider grid for Python Code. 

\begin{table}[H]
\centering
\caption{Full $(p,T)$ sweep for the \methodname{} (Small Teacher) condition, CS, 124M student (15 configurations: $p\in\{0.5,0.75,0.9\}\times T\in\{1.25,1.5,1.75,2.0,2.5\}$). Validation loss, rounded to 2 decimal places; the domain's best cell in \textbf{bold} (ties within rounding are all bolded).}
\label{tab:pt-grid-cs}
\begin{tabular}{lccccc}
\toprule
$p \backslash T$ & $1.25$ & $1.5$ & $1.75$ & $2$ & $2.5$ \\
\midrule
$0.5$ & \textbf{2.73} & \textbf{2.73} & \textbf{2.73} & 2.75 & 2.75 \\
$0.75$ & 2.75 & 2.74 & 2.75 & 2.76 & 2.76 \\
$0.9$ & 2.77 & 2.77 & 2.77 & 2.78 & 2.77 \\
\bottomrule
\end{tabular}
\end{table}

\begin{table}[H]
\centering
\caption{Full $(p,T)$ sweep for the \methodname{} (Small Teacher) condition, BioMed, 124M student (15 configurations: $p\in\{0.5,0.75,0.9\}\times T\in\{1.25,1.5,1.75,2.0,2.5\}$). Validation loss, rounded to 2 decimal places; the domain's best cell in \textbf{bold} (ties within rounding are all bolded).}
\label{tab:pt-grid-biomed}
\begin{tabular}{lccccc}
\toprule
$p \backslash T$ & $1.25$ & $1.5$ & $1.75$ & $2$ & $2.5$ \\
\midrule
$0.5$ & 2.50 & 2.49 & 2.49 & \textbf{2.47} & 2.49 \\
$0.75$ & 2.51 & 2.50 & 2.50 & 2.50 & 2.49 \\
$0.9$ & 2.51 & 2.50 & 2.51 & 2.50 & 2.50 \\
\bottomrule
\end{tabular}
\end{table}

\begin{table}[H]
\centering
\caption{Full $(p,T)$ sweep for the \methodname{} (Small Teacher) condition, News, 124M student (15 configurations: $p\in\{0.5,0.75,0.9\}\times T\in\{1.25,1.5,1.75,2.0,2.5\}$). Validation loss, rounded to 2 decimal places; the domain's best cell in \textbf{bold} (ties within rounding are all bolded).}
\label{tab:pt-grid-news}
\begin{tabular}{lccccc}
\toprule
$p \backslash T$ & $1.25$ & $1.5$ & $1.75$ & $2$ & $2.5$ \\
\midrule
$0.5$ & 2.25 & 2.24 & \textbf{2.23} & 2.25 & 2.26 \\
$0.75$ & 2.26 & 2.28 & 2.29 & 2.28 & 2.30 \\
$0.9$ & 2.33 & 2.33 & 2.33 & 2.33 & 2.33 \\
\bottomrule
\end{tabular}
\end{table}

\begin{table}[H]
\centering
\caption{Full $(p,T)$ sweep for the \methodname{} (Small Teacher) condition, Reviews, 124M student (15 configurations: $p\in\{0.5,0.75,0.9\}\times T\in\{1.25,1.5,1.75,2.0,2.5\}$). Validation loss, rounded to 2 decimal places; the domain's best cell in \textbf{bold} (ties within rounding are all bolded).}
\label{tab:pt-grid-reviews}
\begin{tabular}{lccccc}
\toprule
$p \backslash T$ & $1.25$ & $1.5$ & $1.75$ & $2$ & $2.5$ \\
\midrule
$0.5$ & 3.16 & \textbf{3.15} & \textbf{3.15} & \textbf{3.15} & 3.16 \\
$0.75$ & 3.16 & 3.16 & 3.16 & 3.17 & 3.18 \\
$0.9$ & 3.18 & 3.18 & 3.18 & 3.18 & 3.18 \\
\bottomrule
\end{tabular}
\end{table}

\begin{table}[H]
\centering
\caption{Full $(p,T)$ sweep for the \methodname{} (Small Teacher) condition, Python Code, 124M student. Validation loss, rounded to 2 decimal places; the domain's best cell in \textbf{bold} (ties within rounding are all bolded). The high-temperature extension $T\in\{2.25,2.5,3\}$ was run only for the two competitive mixing ratios, so $p\in\{0.25,0.9\}$, which are uniformly worse at every $T$ that was run ($1.08$ and $0.88$ at best), are omitted.}
\label{tab:pt-grid-pythoncode}
\begin{tabular}{lcccccccc}
\toprule
$p \backslash T$ & $1$ & $1.25$ & $1.5$ & $1.75$ & $2$ & $2.25$ & $2.5$ & $3$ \\
\midrule
$0.5$ & 1.02 & 0.94 & 0.91 & 0.88 & 0.87 & \textbf{0.86} & \textbf{0.86} & 0.87 \\
$0.75$ & 0.93 & 0.91 & 0.89 & 0.88 & 0.88 & 0.87 & 0.88 & 0.88 \\
\bottomrule
\end{tabular}
\end{table}
\begin{table}[H]
\centering
\caption{Optimal temperature $T$ per domain for the 124M student, both teacher conditions. The optimal mixing ratio was $p=0.5$ in every one of the 10 (domain $\times$ teacher) cells tested -- omitted from the table since it does not vary.}
\label{tab:optimal-pt-small-student}
\begin{tabular}{lcc}
\toprule
Domain & Small Teacher: best $T$ & Large Teacher: best $T$ \\
\midrule
Python Code & 2.5 & 2.5 \\
CS          & 1.5 & 1.0 \\
BioMed      & 2.0 & 2.0 \\
News        & 1.75 & 1.5 \\
Reviews     & 1.75 & 1.0 \\
\bottomrule
\end{tabular}
\end{table}

Table~\ref{tab:optimal-pt-large-student} reports the analogous optimal $(p,T)$ for the 1.5B student, \methodname's self-teacher condition (Table~\ref{tab:domain-results-large} and Table~\ref{tab:teacher-strength-ablation}'s ``1.5B \& 1.5B (self)'' row) -- the only 1.5B-student condition that received a real per-domain $(p,T)$ search in this work. Unlike the 124M student, the optimal $p$ is not uniform across domains, so both $p$ and $T$ are reported. We do not report an analogous table for the 1.5B student paired with the 124M teacher (Table~\ref{tab:teacher-strength-ablation}'s ``1.5B \& 124M'' row): that condition reuses the $(p,T)$ found optimal at the 124M-student scale rather than being independently retuned, so there is no separate search to report.

\begin{table}[H]
\centering
\caption{Optimal $(p,T)$ per domain for the 1.5B student, \methodname{} self-teacher condition (1.5B teacher).}
\label{tab:optimal-pt-large-student}
\begin{tabular}{lcc}
\toprule
Domain & Best $p$ & Best $T$ \\
\midrule
Python Code & 0.5 & 1.5 \\
CS & 0.75 & 2.0 \\
BioMed      & 0.5 & 1.0 \\
News        & 0.5 & 1.5 \\
Reviews     & 0.9 & 2.5 \\
\bottomrule
\end{tabular}
\end{table}

\subsection{Instruction Tuning}
\label{sec:appendix:details-instruct}

\subsubsection{Training hyperparameters}
\label{sec:appendix:instruct-hparams}

Every instruction-tuning run in this paper uses the same training recipe, listed in Table~\ref{tab:instruct-hparams}; vanilla SFT and \methodname{} differ only in the loss. Runs are trained with the OLMo-core trainer, starting from each model's released base checkpoint. The recipe follows Ai2's instruction-tuning recipe for OLMo~3 (learning rate, schedule, batch size and sequence length) and is kept fixed across base models and model sizes; we did not tune it per model. The one exception is Llama-3.1-8B (Table~\ref{tab:instruct-families}), for which both vanilla SFT and \methodname{} use a peak learning rate of $2\times10^{-5}$. The only settings that change between experiments are the number of epochs (Table~\ref{tab:instruct-epochs}) and the tokenizer, which is each base model's own. Every run uses 16 GH200 GPUs; the global batch is fixed in tokens, so the gradient-accumulation factor absorbs the context-parallel degree.

\begin{table}[H]
\centering
\caption{Training hyperparameters shared by all instruction-tuning runs.}
\label{tab:instruct-hparams}
\footnotesize
\begin{tabular}{p{0.28\linewidth}p{0.66\linewidth}}
\toprule
Setting & Value \\
\midrule
\multicolumn{2}{l}{\textit{Optimization (vanilla SFT and \methodname{})}} \\
Optimizer & AdamW, $\beta_1=0.9$, $\beta_2=0.95$, $\epsilon=10^{-8}$, weight decay 0 \\
Peak learning rate & $8\times10^{-5}$ \\
Schedule & linear warmup over the first $3\%$ of steps, then linear decay to 0 \\
Gradient clipping & global norm 1.0 \\
Global batch & 1{,}048{,}576 tokens (32 sequences of 32{,}768 tokens) \\
Sequence length & 32{,}768 tokens; documents packed, longer documents truncated \\
Loss tokens & assistant turns only (all other tokens are masked out of the loss) \\
Precision & bfloat16 parameters, float32 gradient reduction \\
Parallelism & FSDP, context parallelism of degree 2 or 4, full activation checkpointing \\
Seeds & data order 34521 \\
\midrule
\multicolumn{2}{l}{\textit{\methodname{} only}} \\
Clean-loss probability $p_c$ & 0.75 (the loss is chosen per micro-batch) \\
Temperature $T$ & 2.5 \\
Teacher & the frozen base checkpoint the student starts from \\
Matching loss & cross-entropy to the teacher's temperature-$T$ distribution over the full vocabulary on the same loss tokens \\
\bottomrule
\end{tabular}
\end{table}

\begin{table}[H]
\centering
\caption{Number of epochs and optimizer steps per experiment. Steps follow from the data size and the fixed global batch.}
\label{tab:instruct-epochs}
\footnotesize
\resizebox{\linewidth}{!}{%
\begin{tabular}{llcc}
\toprule
Experiment & Data & Epochs & Steps \\
\midrule
Tables~\ref{tab:instruct-lowdata}--\ref{tab:instruct-families} & $1\%$ (21{,}503 examples) & 10 & 170 \\
Figure~\ref{fig:epoch-sweep} & $10\%$ (215{,}027 examples) & 2, 4, 6, 8, 10 & 340 to 1{,}700 \\
Figure~\ref{fig:data-scaling}, $1\%$ & $1\%$ & 10 (both methods) & 170 \\
Figure~\ref{fig:data-scaling}, $10\%$ & $10\%$ & 2 (vanilla SFT), 4 (\methodname{}) & 340, 680 \\
Figure~\ref{fig:data-scaling}, $100\%$ & $100\%$ (2{,}150{,}273 examples) & 2 (both methods) & 3{,}376 \\
\bottomrule
\end{tabular}}
\end{table}

\subsubsection{Evaluation harness}
\label{sec:appendix:harness}

All instruction-tuning results (Tables~\ref{tab:instruct-lowdata}, \ref{tab:instruct-families} and Figure~\ref{fig:epoch-sweep}) come from a single, fixed evaluation setup, applied identically to every checkpoint. We use the \texttt{olmes} evaluation suite~\citep{gu2406olmes} (\texttt{github.com/allenai/olmes}, commit \texttt{5a51f502}, unmodified) with the vLLM backend (vLLM 0.13.0, PyTorch 2.9.0, Transformers 4.57.6, Python 3.11). Trained checkpoints are converted to the Hugging Face format and evaluated with the chat template stored alongside them: the OLMo~3 instruct template (\texttt{<|im\_start|>}/\texttt{<|im\_end|>} turns with its default system prompt), which is also the template used to format the training data, for all three base models. Every checkpoint is evaluated once per task, with no selection over seeds or decoding settings.

Table~\ref{tab:harness-tasks} lists the eight task specifications. ``Chat'' tasks format each prompt (including the few-shot examples, as earlier turns) with the chat template; ``plain'' tasks feed a raw text prompt. The reported score is the task's primary metric; for MMLU, MATH and BBH it is the unweighted mean over their subjects or subtasks. The relative change reported in the last column of the instruction-tuning tables is the mean over the eight tasks of $(\text{score}-\text{score}_{\text{SFT}})/\text{score}_{\text{SFT}}$.

\begin{table}[H]
\centering
\caption{Task specifications used for every instruction-tuning evaluation. All generation is greedy except HumanEval+, which draws 20 samples per problem to estimate pass@10. MATH uses the Minerva chain-of-thought prompt format~\citep{lewkowycz2022solving}.}
\label{tab:harness-tasks}
\footnotesize
\setlength{\tabcolsep}{4pt}
\resizebox{\linewidth}{!}{%
\begin{tabular}{llllll}
\toprule
Benchmark & \texttt{olmes} task & Shots & Prompt & Decoding & Metric \\
\midrule
MATH & \texttt{minerva\_math::tulu} & 4, CoT & chat & greedy, $\le$1024 tokens & exact match, mean of 7 subjects \\
GSM8K & \texttt{gsm8k::olmes} & 8, CoT & plain & greedy, $\le$512 tokens & exact match \\
HumanEval+ & \texttt{codex\_humanevalplus::tulu} & 0 & chat & 20 samples, $T{=}0.8$, top-$p$ 0.95 & pass@10 \\
MMLU & \texttt{mmlu::olmes} & 5 & plain & log-likelihood over options & accuracy, mean of 57 subjects \\
PopQA & \texttt{popqa::tulu} & 15 & chat & greedy, $\le$15 tokens & exact match \\
TriviaQA & \texttt{triviaqa::olmes} & 5 & plain & greedy, $\le$50 tokens & F1 (closed-book) \\
IFEval & \texttt{ifeval::tulu} & 0 & chat & greedy, $\le$2048 tokens & prompt-level loose accuracy \\
BBH & \texttt{bbh:cot::olmes} & 3, CoT & plain & greedy, $\le$512 tokens & exact match, mean of 27 tasks \\
\bottomrule
\end{tabular}}
\end{table}

\noindent \textbf{Commands.} The environment and the evaluation of one checkpoint \texttt{\$CKPT} (a Hugging Face model directory) are:
{\footnotesize
\begin{verbatim}
git clone https://github.com/allenai/olmes.git && cd olmes
git checkout 5a51f502d463b8cdc4a2dcad7d7096c41ff1197e
pip install -e . && pip install vllm==0.13.0

export VLLM_WORKER_MULTIPROC_METHOD=spawn
MARGS="gpu_memory_utilization=0.85,max_length=32768"
MARGS="$MARGS,trust_remote_code=false,add_bos_token=false"
TASKS="ifeval::tulu mmlu::olmes gsm8k::olmes codex_humanevalplus::tulu"
TASKS="$TASKS minerva_math::tulu popqa::tulu"
TASKS="$TASKS triviaqa::olmes bbh:cot::olmes"
for TASK in $TASKS; do
  olmes --model-type vllm --model "$CKPT" --model-args "$MARGS" \
        --task "$TASK" --output-dir "$OUT/$(echo $TASK | tr ':/' '__')"
done
\end{verbatim}
}
\noindent For Llama-3.1-8B we set \texttt{add\_bos\_token=true}: \texttt{olmes} encodes prompts without special tokens, and a Llama model trained with a beginning-of-sequence token on every document produces degenerate output without it.

\noindent \textbf{Sharding.} To finish each evaluation in minutes rather than hours, we split every task's instances into 16 disjoint shards and run the command above on each shard in parallel, one GPU per shard. Per-subtask scores are combined across shards weighted by instance count, and the subject and subtask means are taken afterwards, which reproduces the aggregation of an unsharded run.

\subsection{Reasoning}
\label{sec:appendix:details-reasoning}

\subsubsection{Evaluation harness}
\label{sec:appendix:harness-reasoning}

Reasoning (``Think'') checkpoints are evaluated with the same software as Appendix~\ref{sec:appendix:harness} (\texttt{olmes} commit \texttt{5a51f502}, vLLM 0.13.0) but with the \texttt{olmo3:adapt} task specifications, the ones behind Ai2's OLMo~3 Think model card. Every task is zero-shot and chat-formatted with the OLMo~3 Think chat template, whose generation prompt ends with \texttt{<|im\_start|>assistant\textbackslash n<think>}, so every response starts inside a reasoning block and the model must close it with \texttt{</think>} before its final answer. All generation is sampled with temperature 0.6 and top-$p$ 0.95. The model context is capped at 32{,}768 tokens (prompt plus response); AIME additionally caps the response at 16{,}384 tokens. Table~\ref{tab:harness-reasoning-tasks} lists the six tasks.

\begin{table}[H]
\centering
\caption{Task specifications used for every reasoning evaluation. All tasks are zero-shot, chat-formatted and sampled at $T{=}0.6$, top-$p$ 0.95.}
\label{tab:harness-reasoning-tasks}
\footnotesize
\setlength{\tabcolsep}{4pt}
\resizebox{\linewidth}{!}{%
\begin{tabular}{lllll}
\toprule
Benchmark & \texttt{olmes} task & Samples per problem & Response budget & Metric \\
\midrule
GSM8K & \texttt{gsm8k::olmo3:adapt} & 1 & context limit & exact match of the \texttt{\textbackslash boxed\{\}} answer \\
AIME 2024 & \texttt{aime:2024::olmo3:adapt} & 32 & 16{,}384 tokens & pass@1, mean over the 32 samples \\
AIME 2025 & \texttt{aime:2025::olmo3:adapt} & 32 & 16{,}384 tokens & pass@1, mean over the 32 samples \\
HumanEval+ & \texttt{codex\_humanevalplus::olmo3:adapt} & 10 & context limit & pass@1 \\
IFEval & \texttt{ifeval::olmo3:adapt} & 1 & context limit & prompt-level loose accuracy \\
GPQA & \texttt{gpqa::olmo3:adapt} & 1 & context limit & exact match of the answer letter (448 questions) \\
\bottomrule
\end{tabular}}
\end{table}

\noindent \textbf{Commands.} With the environment of Appendix~\ref{sec:appendix:harness}, one Think checkpoint \texttt{\$CKPT} is evaluated with:
{\footnotesize
\begin{verbatim}
export VLLM_WORKER_MULTIPROC_METHOD=spawn
MARGS="gpu_memory_utilization=0.85,max_length=32768"
MARGS="$MARGS,trust_remote_code=false,add_bos_token=false"
TASKS="gsm8k::olmo3:adapt aime:2024::olmo3:adapt aime:2025::olmo3:adapt"
TASKS="$TASKS ifeval::olmo3:adapt codex_humanevalplus::olmo3:adapt"
TASKS="$TASKS gpqa::olmo3:adapt"
for TASK in $TASKS; do
  olmes --model-type vllm --model "$CKPT" --model-args "$MARGS" \
        --task "$TASK" --output-dir "$OUT/$(echo $TASK | tr ':/' '__')"
done
\end{verbatim}
}
\noindent The same 16-way sharding and instance-weighted aggregation as in Appendix~\ref{sec:appendix:harness} are used. Because generation is sampled, repeated evaluations of the same checkpoint differ slightly; AIME's 32 samples per problem keep the sampling standard error of its score at about one point (0.8 to 1.2 points for the checkpoints in this paper).

\section{Additional Results}
\label{sec:appendix:additional-results}

\subsection{Full temperature sweep on Tiny Shakespeare}
\label{sec:appendix:shakespeare-full-sweep}

\begin{figure}[h]
    \centering
    \includegraphics[width=0.62\linewidth]{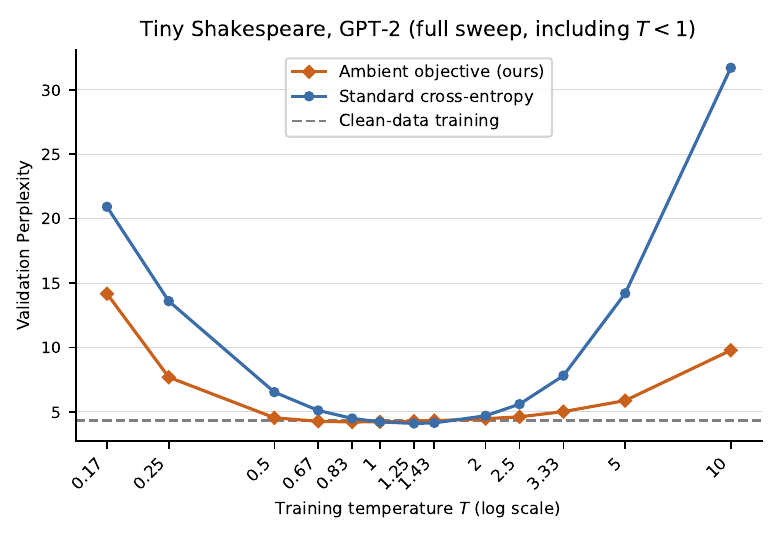}
    \caption{The sweep of Figure~\ref{fig:shakespeare_dataset} over the full range of training temperatures, $T=1/6$ to $T=10$. Both losses degrade at both extremes; the objective of Eq.~\eqref{eq:ambient_loss} degrades far more slowly.}
    \label{fig:shakespeare_full_sweep}
\end{figure}

Figure~\ref{fig:shakespeare_dataset} shows the $T\ge1$ half of the sweep, the regime the method operates in. Figure~\ref{fig:shakespeare_full_sweep} adds the low-temperature half, which is symmetric: sampling far below $T=1$ is also a corruption, producing repetitive rather than random text, and the objective of Eq.~\eqref{eq:ambient_loss} again tracks clean-data training over a much wider range of $T$ than standard cross-entropy does.

\subsection{Domain Adaptation}
\label{sec:appendix:additional-domain}

\subsubsection{Predicting the success of \methodname}
\label{sec:appendix:zeroshot-perplexity}

Table~\ref{tab:zeroshot-domain-perplexity} reports the zero-shot validation loss of the pre-trained (1.5B) model on each target domain, before any fine-tuning, alongside \methodname's win margin over the best baseline from Table~\ref{tab:domain-results-large}, as discussed in Section~\ref{sec:domain-shift-experiments}.

\begin{table}[H]
\centering
\caption{Zero-shot validation loss of the pre-trained (1.5B) model on each target domain, before any fine-tuning, alongside \methodname's win margin over the best baseline from Table~\ref{tab:domain-results-large} (best baseline loss minus \methodname{} loss; higher is a bigger win; negative means \methodname{} was outperformed). Domains are ordered by zero-shot loss.}
\label{tab:zeroshot-domain-perplexity}
\begin{tabular}{lccccc}
\toprule
 & Python Code & News & BioMed & CS & Reviews \\
\midrule
Zero-shot validation loss & 1.46 & 2.59 & 2.61 & 2.87 & 3.56 \\
\methodname{} win margin & 0.09 & 0.31 & 0.01 & 0.01 & 0.00 \\
\bottomrule
\end{tabular}
\end{table}

\subsubsection{Ablation: effect of a smaller student}
\label{sec:appendix:student-size-ablation}

Table~\ref{tab:domain-results-small} repeats the domain-adaptation comparison of Table~\ref{tab:domain-results-large} with a 124M-parameter student in place of the 1.5B one, holding every other condition fixed (same target domains, same OWT-pretrained 1.5B teacher for \methodname). \textit{\methodname{} (Small Teacher)} additionally reports a self-teacher variant, where the frozen teacher is the same 124M OWT-pretrained checkpoint as the student's own initialization, rather than the larger 1.5B model. Comparing against Table~\ref{tab:domain-results-large} isolates the effect of student capacity: every resume-based method improves substantially with the larger student, and \methodname's margin over the baselines is preserved across the scale change.

\begin{table}[H]
\centering
\caption{Best validation loss achieved per domain across fine-tuning strategies, 124M student. \textit{Target Only (From Scratch)}: trained from scratch on target-domain data only. \textit{Target Only Fine-tuning}: finetuned from the OWT-pretrained checkpoint on target-domain data only. \textit{Data Mixture Fine-tuning}: finetuned from the OWT-pretrained checkpoint on a clean mixture of target-domain and OWT data (best mixing ratio $p$ per domain). \textit{\methodname{} (Small Teacher)}: our method against a frozen self-teacher (the same 124M OWT-pretrained checkpoint), using only target-domain data (best $p, T$ per domain). \textit{\methodname{} (Large Teacher)}: same mechanism, with a frozen 1.5B-parameter teacher in place of the 124M self-teacher. Values rounded to 2 decimal places, the level of precision supported by measured run-to-run variance across random seeds ($\sigma\approx0.003$--$0.006$ across two domains checked); differences below this are not distinguishable from noise. Lower is better; best result(s) per domain in \textbf{bold} (ties within rounding are all bolded).}
\label{tab:domain-results-small}
\begin{tabular}{lccccc}
\toprule
Method & Python Code & CS & BioMed & News & Reviews \\
\midrule
Target Only (From Scratch) & 0.98 & 2.99 & 2.62 & 2.63 & 3.34 \\
Target Only Fine-tuning     & 0.88 & 2.78 & 2.51 & 2.32 & 3.19 \\
Data Mixture Fine-tuning    & 0.88 & 2.69 & 2.47 & 2.26 & \textbf{3.11} \\
\hline
\methodname{} (Small Teacher)    & 0.86 & 2.73 & 2.47 & \textbf{2.23} & 3.15 \\
\methodname{} (Large Teacher)    & \textbf{0.83} & \textbf{2.68} & \textbf{2.43} & \textbf{2.23} & 3.12 \\
\bottomrule
\end{tabular}
\end{table}

\subsubsection{Ablation: effect of a weaker teacher}
\label{sec:appendix:teacher-strength-ablation}

Table~\ref{tab:teacher-strength-ablation} isolates the effect of teacher strength on \methodname, holding the student fixed and varying only whether the frozen teacher is the small (124M) or large (1.5B) OWT-pretrained model. We report this comparison at both student scales considered in this work. At both scales, using the larger teacher matches or improves over the smaller one on every domain, indicating that our method continues to benefit from a stronger teacher even though, unlike classical KD, the teacher is never the stronger model at the target task -- it is only used as a coarse, high-temperature regularizer (Section~\ref{sec:proxy}).

\begin{table}[H]
\centering
\caption{Effect of teacher strength on \methodname's best validation loss, at both student scales. The 124M teacher is a self-teacher when the student is also 124M; the 1.5B teacher is the same OWT-pretrained checkpoint used as the teacher in Table~\ref{tab:domain-results-large}. Values rounded to 2 decimal places; best teacher choice per domain (within each student-size block) in \textbf{bold} (ties within rounding are all bolded).}
\label{tab:teacher-strength-ablation}
\begin{tabular}{llccccc}
\toprule
Student & Teacher & Python Code & CS & BioMed & News & Reviews \\
\midrule
124M & 124M (self) & 0.86 & 2.73 & 2.47 & \textbf{2.23} & 3.15 \\
124M & 1.5B         & \textbf{0.83} & \textbf{2.68} & \textbf{2.43} & \textbf{2.23} & \textbf{3.12} \\
\hline
1.5B & 124M         & 0.79 & 2.61 & 2.39 & 1.96 & 3.01 \\
1.5B & 1.5B (self)  & \textbf{0.70} & \textbf{2.54} & \textbf{2.35} & \textbf{1.68} & \textbf{2.96} \\
\bottomrule
\end{tabular}
\end{table}

\subsection{Instruction Tuning}
\label{sec:appendix:instruct-extra}

\begin{figure}[h]
\centering
\includegraphics[width=\linewidth]{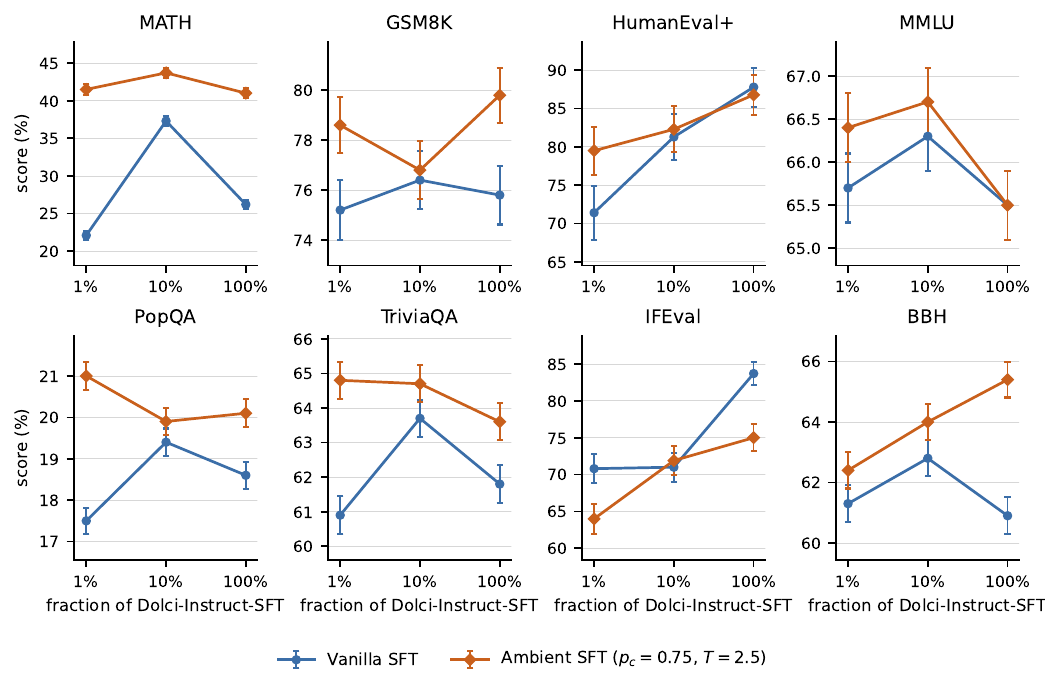}
\caption{Benchmark scores versus the fraction of Dolci-Instruct-SFT data used for training. \small{For all these results, we start from an OLMo-3-7B model. Error bars are $\pm1$ binomial standard error over each task's items. For the majority of the benchmarks, \methodname{} improves upon regular SFT across all data scales. The highest improvements are observed in the $1\%$ of the dataset setting.}}
\label{fig:data-scaling}
\end{figure}

\begin{figure}[h]
\centering
\includegraphics[width=\linewidth]{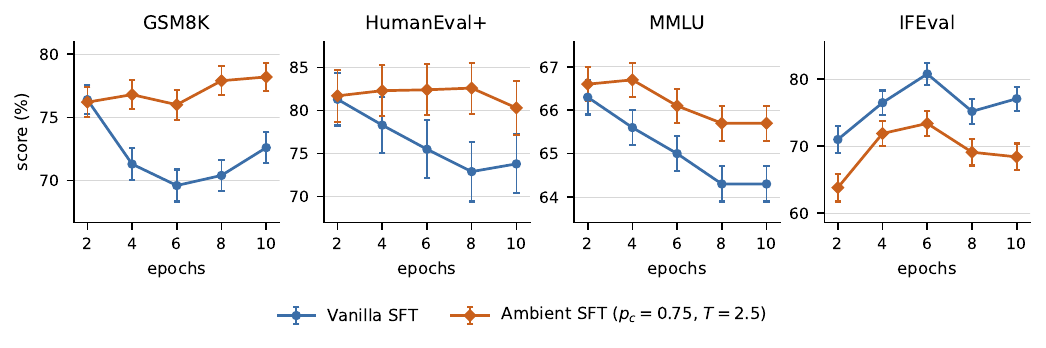}
\caption{The remaining four tasks of Figure~\ref{fig:epoch-sweep}: benchmark scores versus the number of epochs on a $10\%$ subset of Dolci-Instruct-SFT, OLMo-3-7B, error bars $\pm1$ binomial standard error.}
\label{fig:epoch-sweep-appendix}
\end{figure}

Figure~\ref{fig:data-scaling} reports the full data-availability sweep discussed in Section~\ref{sec:instruct-experiments}, and Figure~\ref{fig:epoch-sweep-appendix} completes Figure~\ref{fig:epoch-sweep} with the four tasks left out of the main paper.

\label{sec:appendix:additional-instruct}

\subsection{Baselines for instruction tuning}
\label{sec:appendix:baselines}
We compare \methodname{} with four baselines, each swept over a grid whose best value lies inside the grid:
\begin{compactitem}
\item \textbf{L2-SP}~\citep{xuhong2018explicit} penalizes the distance of the weights from the pre-trained weights. We sweep the penalty strength $\alpha\in\{0.001, 0.01, 0.1, 1, 10\}$.
\item \textbf{NEFTune}~\citep{jain2310neftune} adds uniform noise to the input embeddings during fine-tuning. We sweep the noise scale $\alpha\in\{5, 10, 15\}$.
\item \textbf{Knowledge distillation (KD)} adds to vanilla SFT a KL term to the frozen base model at temperature $T=1$, i.e., our objective with $T=1$. We sweep the cross-entropy weight $p_c\in\{0.25, 0.5, 0.75, 0.9\}$.
\item \textbf{Label smoothing} mixes the one-hot target with the uniform distribution with weight $\epsilon$. We sweep $\epsilon\in\{0.05, 0.1, 0.2, 0.4\}$.
\end{compactitem}
The best settings are $\alpha=10$ for NEFTune, $\alpha=1$ for L2-SP, $p_c=0.5$ for KD and $\epsilon=0.2$ for label smoothing. Table~\ref{tab:baselines-full} reports the full sweeps.
\begin{table}[H]
\centering
\caption{Full hyperparameter sweeps of the baselines of Table~\ref{tab:baselines}, Qwen3-1.7B-Base, Table~\ref{tab:instruct-lowdata} setup.}
\label{tab:baselines-full}
\footnotesize
\setlength{\tabcolsep}{3.5pt}
\resizebox{\linewidth}{!}{%
\begin{tabular}{lcccccccc|c}
\toprule
& MATH & GSM8K & \makecell{HumanEval+} & MMLU & PopQA & TriviaQA & IFEval & BBH & \makecell{Avg.\ rel.\ $\Delta$\\vs.\ SFT} \\
\midrule
Base (no SFT) & 33.8 & 53.5 & 78.4 & \textbf{65.6} & 13.9 & \textbf{45.6} & 37.9 & 54.2 & $+9.0\%$ \\
\midrule
Vanilla SFT & 24.3 & 61.0 & 68.0 & 61.9 & 12.2 & 36.3 & 58.8 & 45.2 & --- \\
NEFTune ($\alpha=5$) & 23.7 & 60.0 & 66.8 & 61.6 & 11.7 & 37.8 & 60.8 & 46.8 & $+0.1\%$ \\
NEFTune ($\alpha=10$) & 19.7 & 58.9 & 69.2 & 62.2 & 12.4 & 38.4 & \textbf{65.6} & 48.6 & $+0.8\%$ \\
NEFTune ($\alpha=15$) & 18.3 & 61.9 & 71.7 & 62.2 & 11.8 & 39.0 & 63.0 & 48.7 & $+0.2\%$ \\
L2-SP ($\alpha=0.001$) & 20.2 & 56.0 & 72.5 & 61.4 & 12.2 & 37.1 & 58.4 & 45.8 & $-2.1\%$ \\
L2-SP ($\alpha=0.01$) & 21.6 & 57.9 & \textbf{84.5} & 63.8 & 15.3 & 43.1 & 54.9 & 49.0 & $+7.1\%$ \\
L2-SP ($\alpha=0.1$) & 30.3 & 72.6 & 83.0 & 65.0 & 14.8 & 45.3 & 50.1 & 52.7 & $+14.8\%$ \\
L2-SP ($\alpha=1$) & 35.0 & 72.3 & 83.2 & 64.9 & 14.5 & \textbf{45.6} & 41.8 & 52.8 & $+15.3\%$ \\
L2-SP ($\alpha=10$) & 35.1 & 69.4 & 81.8 & 65.2 & 13.9 & \textbf{45.6} & 38.8 & 53.7 & $+13.5\%$ \\
KD ($T=1$, $p_c=0.25$) & 31.8 & \textbf{72.9} & 80.5 & 65.0 & 15.6 & 44.4 & 43.1 & \textbf{55.2} & $+14.9\%$ \\
KD ($T=1$, $p_c=0.5$) & 35.1 & 71.5 & 81.3 & 64.8 & \textbf{15.7} & 43.8 & 49.2 & 52.8 & $+17.0\%$ \\
KD ($T=1$, $p_c=0.75$) & 37.5 & 67.9 & 79.0 & 64.3 & 13.6 & 42.6 & 55.5 & 52.4 & $+15.6\%$ \\
KD ($T=1$, $p_c=0.9$) & 32.4 & 63.8 & 78.4 & 62.9 & 13.3 & 40.9 & 61.9 & 51.7 & $+12.0\%$ \\
Label smoothing ($\epsilon=0.05$) & 19.4 & 57.9 & 68.8 & 61.1 & 12.0 & 36.0 & 59.7 & 47.5 & $-2.7\%$ \\
Label smoothing ($\epsilon=0.1$) & 25.3 & 61.4 & 64.2 & 62.3 & 10.0 & 37.3 & 57.7 & 44.7 & $-2.3\%$ \\
Label smoothing ($\epsilon=0.2$) & 25.4 & 62.9 & 64.3 & 60.9 & 12.1 & 36.4 & 55.1 & 47.0 & $-0.3\%$ \\
Label smoothing ($\epsilon=0.4$) & 23.5 & 62.5 & 56.8 & 61.0 & 12.0 & 34.9 & 54.2 & 46.9 & $-3.5\%$ \\
\midrule
\methodname{} & \textbf{40.4} & 72.3 & 83.7 & 64.9 & 14.9 & 43.8 & 52.3 & 53.3 & $+20.3\%$ \\
\bottomrule
\end{tabular}}
\end{table}

\subsection{Per-source temperature}
\label{sec:appendix:per-source-T}
Dolci-Instruct-SFT is a mixture of 21 source datasets. For each source $s$, we run the base model once over that source's supervised tokens and measure the gap $g_s$ between its cross-entropy on the observed tokens and its average predictive entropy. We then raise the temperature where the gap is large, so that the student relies less on the base model there. We use the rule
\[
T_s = 2.5\,\left(1+\frac{\max(g_s,0)}{0.1}\right),
\]
which keeps $T=2.5$ for sources the base model fits well. Sources such as math and code get temperatures between $2.5$ and $3.4$, while poorly fit ones, such as CoCoNot and WildChat, get up to $10.7$. This rule is a simple heuristic rather than a tuned choice: we compared three such rules at Qwen3-1.7B, kept this one, and used its constants unchanged at 4B and 8B. Other rules based on the same gap may work better.

\subsection{Reasoning}
\label{sec:appendix:additional-reasoning}
Table~\ref{tab:reasoning-knowledge} evaluates the Vanilla SFT and \methodname{} reasoning checkpoints of Table~\ref{tab:reasoning-baselines} on knowledge benchmarks with the few-shot, non-chat specifications of the instruction-tuning harness (Table~\ref{tab:harness-tasks}) on the full test sets. \methodname{} is ahead on all four benchmarks.

\begin{table}[H]
\centering
\caption{Knowledge benchmarks for the Vanilla SFT and \methodname{} reasoning checkpoints of Table~\ref{tab:reasoning-baselines} (OLMo-3-7B, $1\%$ of Dolci-Think-SFT, ten epochs; \methodname{} with $p_c=0.75$, $T=2.5$), with few-shot, non-chat prompting on the full test sets.}
\label{tab:reasoning-knowledge}
\footnotesize
\begin{tabular}{lcc}
\toprule
Benchmark & Vanilla SFT & \methodname{} \\
\midrule
MMLU & 64.6 & \textbf{66.0} \\
GSM8K & 77.6 & \textbf{83.7} \\
TriviaQA & 57.0 & \textbf{64.3} \\
BBH & 57.0 & \textbf{66.2} \\
\bottomrule
\end{tabular}
\end{table}

\section{Theory}
\subsection{Proof of Theorem \ref{th:learning_from_high_temperature}}
\label{sec:appendix:high-temp-proof}

\begin{proof}
\emph{(a)} For distributions $q,r$ on $V$ with $r$ of full support, $D_{\mathrm{KL}}(q\|r)=\sum_y q(y)\log q(y)-\sum_y q(y)\log r(y)=\mathbb E_{y\sim q}[-\log r(y)]-H(q)$. For any $\vx_{<i}$, we can plug into this identity $q=p_{T\theta^*}(\cdot\mid\vx_{<i})$ and $r=p_{T\theta}(\cdot\mid\vx_{<i})$, to get the following:
\begin{align}
D_{\mathrm{KL}}(p_{T\theta^*}(\cdot\mid\vx_{<i})\|p_{T\theta}(\cdot\mid\vx_{<i}))=\mathbb E_{x_i \sim p_{T\theta^*}(\cdot\mid\vx_{<i})}\big[ -\log p_{T\theta}(x_i\mid\vx_{<i})  \big] - H(p_{T\theta^*}(\cdot\mid\vx_{<i})). \label{eq:cos 1}
\end{align}
Taking the expectation of~\eqref{eq:cos 1} over $\vx_{<i}\sim\mu_i$ and summing over $i$:
\begin{align*}
J_T(\theta)&=\sum_{i=1}^{l}\ \mathbb E_{\vx_{<i}\sim \mu_i}\ \mathbb E_{x_i\sim p_{T\theta^*}(\cdot\mid\vx_{<i})}\big[-\log p_{T\theta}(x_i\mid\vx_{<i})\big]\\
&=\sum_{i=1}^{l}\ \mathbb E_{\vx_{<i}\sim \mu_i}\Big[D_{\mathrm{KL}}\big(p_{T\theta^*}(\cdot\mid\vx_{<i})\,\big\|\,p_{T\theta}(\cdot\mid\vx_{<i})\big)\Big]+\mathrm{const},
\end{align*}
i.e., $J_T(\theta)=R_T(\theta)+\mathrm{const}$, where the constant is the $\mu$-weighted sum of the teacher's tempered entropies, which depends on $\theta^*$ and $\mu$ but not $\theta$. The argument never uses where the prefixes come from.

\emph{(b)} For all $\theta$, we have that: $J_T(\theta)-J_T(\theta^*)=\sum_i\mathbb E_{\vx_{<i}\sim \mu_i}[D_{\mathrm{KL}}(p_{T\theta^*}(\cdot\mid\vx_{<i})\|p_{T\theta}(\cdot\mid\vx_{<i}))]\ge0$. So $\theta^*$ is clearly a minimizer. Moreover, $\theta$ is a minimizer iff every KL term with positive weight vanishes, i.e., iff $p_{T\theta}(\cdot\mid\vx_{<i})=p_{T\theta^*}(\cdot\mid\vx_{<i})$ for every context $\vx_{<i}$ in the support of $\mu_i$ (the set $V^{i-1}$ is finite, so these are the contexts with $\mu_i(\vx_{<i})>0$). On such a context, this holds iff $g_\theta(y\mid\vx_{<i})/T-g_{\theta^*}(y\mid\vx_{<i})/T$ is a constant function of $y\in V$, i.e., iff $g_\theta(\cdot\mid\vx_{<i})/T=g_{\theta^*}(\cdot\mid\vx_{<i})/T+c(\vx_{<i})$, i.e.~iff $g_\theta(\cdot\mid\vx_{<i})/T'=g_{\theta^*}(\cdot\mid\vx_{<i})/T'+c(\vx_{<i})$ for all $T'>0$, i.e.~iff $p_{T'\theta}(\cdot\mid\vx_{<i})=p_{T'\theta^*}(\cdot\mid\vx_{<i})$ for every $T'>0$. Contexts outside the support of $\mu_i$ do not enter $J_T$ and are not constrained.
\end{proof}

\begin{corollary}[Prefixes sampled from the tempered teacher]
\label{cor:tempered-prefixes}
Let $p_{T\theta^*}(\vx)=\prod_{i=1}^{l}p_{T\theta^*}(x_i\mid\vx_{<i})$ be the distribution of text sampled autoregressively at temperature $T$, and let $\mu_i$ be the distribution of its first $i-1$ tokens. Then
\begin{gather}
    J_T(\theta)=\mathbb E_{\vx\sim p_{T\theta^*}}\left[\sum_{i=1}^{l}-\log p_{T\theta}(x_i\mid\vx_{<i})\right]
    \label{eq:ambient_loss}
\end{gather}
is the tempered cross-entropy on text sampled at temperature $T$, every context lies in the support of $\mu_i$, and every minimizer of $J_T$ matches the teacher at all temperatures on every context.
\end{corollary}

\begin{proof}[Proof of Corollary~\ref{cor:tempered-prefixes}]
Since $p_{T\theta^*}(\vx)$ is the autoregressive product of the tempered conditionals, under $\vx\sim p_{T\theta^*}$ the prefix $\vx_{<i}$ has distribution $\mu_i(\vx_{<i})=\prod_{j<i}p_{T\theta^*}(x_j\mid\vx_{<j})$ and, given the prefix, $x_i\sim p_{T\theta^*}(\cdot\mid\vx_{<i})$. By the tower rule, $\mathbb E_{\vx\sim p_{T\theta^*}}[-\log p_{T\theta}(x_i\mid\vx_{<i})]=\mathbb E_{\vx_{<i}\sim\mu_i}\mathbb E_{x_i\sim p_{T\theta^*}(\cdot\mid\vx_{<i})}[-\log p_{T\theta}(x_i\mid\vx_{<i})]$, and summing over $i$ gives Eq.~\eqref{eq:ambient_loss} as the instance of Eq.~\eqref{eq:ambient_loss_general} with these $\mu_i$. Because each tempered conditional is a softmax, $\mu_i(\vx_{<i})>0$ for every $\vx_{<i}\in V^{i-1}$, so every context lies in the support of $\mu_i$, and Theorem~\ref{th:learning_from_high_temperature}(b) gives recovery on every context.
\end{proof}

\subsection{Proof of Theorem \ref{thm:high-temperature-convergence}}
\label{sec:appendix:contraction_proof}

\begin{theorem}[High-temperature conditional KL convergence]
\label{thm:high-temperature-convergence}
Fix a context $x_{<i}$ and let $\Delta(x_{<i})$ be the range of the logit gap between the two models,
\begin{align}
\Delta(x_{<i})&=\max_{y\in\mathcal V}\big[g_{\theta_{\mathrm{post}}^*}(y\mid x_{<i})-g_{\theta_{\mathrm{pre}}^*}(y\mid x_{<i})\big]-\min_{y\in\mathcal V}\big[g_{\theta_{\mathrm{post}}^*}(y\mid x_{<i})-g_{\theta_{\mathrm{pre}}^*}(y\mid x_{<i})\big]\\&=\max_{y\in\mathcal V}\big[g_{\theta_{\mathrm{pre}}^*}(y\mid x_{<i})-g_{\theta_{\mathrm{post}}^*}(y\mid x_{<i})\big]-\min_{y\in\mathcal V}\big[g_{\theta_{\mathrm{pre}}^*}(y\mid x_{<i})-g_{\theta_{\mathrm{post}}^*}(y\mid x_{<i})\big]
\end{align}
For every $T>0$:
\[
D_{\mathrm{KL}}\!\left(p_{T\theta_{\mathrm{post}}^*}(\cdot\mid x_{<i})\,\middle\|\,p_{T\theta_{\mathrm{pre}}^*}(\cdot\mid x_{<i})\right)\le\frac{\Delta(x_{<i})^2}{8T^2}.
\]
The same bound holds with the two models interchanged.
\end{theorem}

 \begin{proof}
The KL divergence of interest is
\[
\begin{aligned}
&D_{\mathrm{KL}}\!\left(
p_{T\theta_{\mathrm{post}}^*}(\cdot\mid x_{<i})
\,\middle\|\,
p_{T\theta_{\mathrm{pre}}^*}(\cdot\mid x_{<i})
\right)
\\
&\qquad=
\mathbb{E}_{Y\sim p_{T\theta_{\mathrm{post}}^*}(\cdot\mid x_{<i})}\left[
\log
\frac{
p_{T\theta_{\mathrm{post}}^*}(Y\mid x_{<i})
}{
p_{T\theta_{\mathrm{pre}}^*}(Y\mid x_{<i})
}
\right].
\end{aligned}
\]

We will write the KL divergence in a form to which Hoeffding's lemma can be applied. Before we do this, let us notice the following easy fact:
\[
\begin{aligned}
\mathbb{E}_{Y\sim p_{T\theta_{\mathrm{post}}^*}(\cdot\mid x_{<i})}\left[
\exp\left(
-\log
\frac{
p_{T\theta_{\mathrm{post}}^*}(Y\mid x_{<i})
}{
p_{T\theta_{\mathrm{pre}}^*}(Y\mid x_{<i})
}
\right)
\right]
&=
\sum_{y\in\mathcal V}
p_{T\theta_{\mathrm{post}}^*}(y\mid x_{<i})
\frac{
p_{T\theta_{\mathrm{pre}}^*}(y\mid x_{<i})
}{
p_{T\theta_{\mathrm{post}}^*}(y\mid x_{<i})
}
\\
&=
\sum_{y\in\mathcal V}
p_{T\theta_{\mathrm{pre}}^*}(y\mid x_{<i})
=1.
\end{aligned}
\]
We claim that this implies the following:
\begin{equation}
\begin{aligned}
&D_{\mathrm{KL}}\!\left(
p_{T\theta_{\mathrm{post}}^*}(\cdot\mid x_{<i})
\,\middle\|\,
p_{T\theta_{\mathrm{pre}}^*}(\cdot\mid x_{<i})
\right)
\\
&=
\log
\mathbb{E}\left[
\exp\left(
-\log
\frac{
p_{T\theta_{\mathrm{post}}^*}(Y\mid x_{<i})
}{
p_{T\theta_{\mathrm{pre}}^*}(Y\mid x_{<i})
}
+
\mathbb{E}\left[
\log
\frac{
p_{T\theta_{\mathrm{post}}^*}(Y\mid x_{<i})
}{
p_{T\theta_{\mathrm{pre}}^*}(Y\mid x_{<i})
}
\right]
\right)
\right].
\end{aligned}
\label{eq:kl-centered-mgf}
\end{equation}
Indeed, the RHS of the above equals:
\begin{align*}
&\log \Bigg(
\mathbb{E}\left[
\exp\left(
-\log
\frac{
p_{T\theta_{\mathrm{post}}^*}(Y\mid x_{<i})
}{
p_{T\theta_{\mathrm{pre}}^*}(Y\mid x_{<i})
}\right)\right] \cdot 
\exp \left(
\mathbb{E}\left[
\log
\frac{
p_{T\theta_{\mathrm{post}}^*}(Y\mid x_{<i})
}{
p_{T\theta_{\mathrm{pre}}^*}(Y\mid x_{<i})
}
\right]
\right)
\Bigg)\\&~~~~~~~~~~~~~~~~~~~~~~~~~~~~~~~~~~~~~~~= \log \Bigg(
1\cdot
\exp \left(
\mathbb{E}\left[
\log
\frac{
p_{T\theta_{\mathrm{post}}^*}(Y\mid x_{<i})
}{
p_{T\theta_{\mathrm{pre}}^*}(Y\mid x_{<i})
}
\right]
\right)
\Bigg)\\&~~~~~~~~~~~~~~~~~~~~~~~~~~~~~~~~~~~~~~~=\mathbb{E}\left[
\log
\frac{
p_{T\theta_{\mathrm{post}}^*}(Y\mid x_{<i})
}{
p_{T\theta_{\mathrm{pre}}^*}(Y\mid x_{<i})
}
\right].
\end{align*}
Recall that Hoeffding's lemma implies that, for any random variable $Z$
supported on an interval $[a,b]$,
\[
\log\mathbb{E}\left[
\exp\left(Z-\mathbb{E}[Z]\right)
\right]
\le \frac{(b-a)^2}{8}.
\]
We apply Hoeffding's lemma to the random variable
\[
Z:=-\log
\frac{
p_{T\theta_{\mathrm{post}}^*}(Y\mid x_{<i})
}{
p_{T\theta_{\mathrm{pre}}^*}(Y\mid x_{<i})
}.
\]
This random variable is bounded because the vocabulary is finite and
all logits are finite. Its centered exponential moment is exactly the
right-hand side of \eqref{eq:kl-centered-mgf}. Therefore,
\begin{equation}
\begin{aligned}
&D_{\mathrm{KL}}\!\left(
p_{T\theta_{\mathrm{post}}^*}(\cdot\mid x_{<i})
\,\middle\|\,
p_{T\theta_{\mathrm{pre}}^*}(\cdot\mid x_{<i})
\right)
\\
&\qquad\le
\frac{1}{8}
\left(
\max_{y\in\mathcal V}
\log
\frac{
p_{T\theta_{\mathrm{post}}^*}(y\mid x_{<i})
}{
p_{T\theta_{\mathrm{pre}}^*}(y\mid x_{<i})
}
-
\min_{y\in\mathcal V}
\log
\frac{
p_{T\theta_{\mathrm{post}}^*}(y\mid x_{<i})
}{
p_{T\theta_{\mathrm{pre}}^*}(y\mid x_{<i})
}
\right)^2,
\end{aligned}
\label{eq:conditional-kl-range}
\end{equation}
where we used the fact that negating a random variable does not change
the length of its range.

By the definition of temperature scaling,
\[
\begin{aligned}
\log
\frac{
p_{T\theta_{\mathrm{post}}^*}(y\mid x_{<i})
}{
p_{T\theta_{\mathrm{pre}}^*}(y\mid x_{<i})
}
&=
\frac{
g_{\theta_{\mathrm{post}}^*}(y\mid x_{<i})
-
g_{\theta_{\mathrm{pre}}^*}(y\mid x_{<i})
}{T}
\\
&\quad+
\log
\frac{
\sum_{z\in\mathcal V}
\exp\!\left(
g_{\theta_{\mathrm{pre}}^*}(z\mid x_{<i})/T
\right)
}{
\sum_{z\in\mathcal V}
\exp\!\left(
g_{\theta_{\mathrm{post}}^*}(z\mid x_{<i})/T
\right)
}.
\end{aligned}
\]
The second term is independent of $y$ and therefore cancels when
taking the difference between the maximum and minimum over tokens.
Consequently,
\[
\begin{aligned}
&
\max_{y\in\mathcal V}
\log
\frac{
p_{T\theta_{\mathrm{post}}^*}(y\mid x_{<i})
}{
p_{T\theta_{\mathrm{pre}}^*}(y\mid x_{<i})
}
-
\min_{y\in\mathcal V}
\log
\frac{
p_{T\theta_{\mathrm{post}}^*}(y\mid x_{<i})
}{
p_{T\theta_{\mathrm{pre}}^*}(y\mid x_{<i})
}
\\
&\qquad=
\frac{\Delta(x_{<i})}{T}.
\end{aligned}
\]
Substituting this identity into
\eqref{eq:conditional-kl-range} gives
\[
D_{\mathrm{KL}}\!\left(
p_{T\theta_{\mathrm{post}}^*}(\cdot\mid x_{<i})
\,\middle\|\,
p_{T\theta_{\mathrm{pre}}^*}(\cdot\mid x_{<i})
\right)
\le
\frac{\Delta(x_{<i})^2}{8T^2}.
\]
\end{proof}

\subsection{Proofs for Section~\ref{sec:learning_tradeoffs} (log-linear model)}
\label{sec:appendix:loglinear}
\paragraph{Notation.} {Starred parameters are minimizers of population objectives while hats mark estimators computed from the $n$ observations. $\widehat\theta_\lambda$ denotes the minimizer of our objective and $\widehat\theta_0$ denotes the clean maximum-likelihood estimator ($\lambda=0$). $\theta^*_{\mathrm{pre}}$ is the teacher and $\theta^*_{\mathrm{post}}$ is the target distribution.}

\paragraph{Regularity conditions for Theorem~\ref{thm:loglinear-bias-variance}.} The minimizer $\widehat\theta_\lambda$ of \eqref{eq:loglinear-objective} is taken over a fixed ball $\mathcal B_R=\{\|A_\theta\|_{\mathrm{op}}\le R\}$ containing $\theta^*_{\mathrm{pre}}$ and $\theta^*_{\mathrm{post}}$, a technical restriction that keeps all logits bounded. The embeddings are bounded, $\|\phi(y)\|\le R_\phi$ for every $y\in V$ and $\|\psi\|\le R_\psi$ almost surely, so all logits on $\mathcal B_R$ are bounded by $\varrho=RR_\phi R_\psi$. We assume that the minimizer is in the interior of the ball. We also recall the two non-degeneracy conditions of Section~\ref{sec:learning_tradeoffs}: $G=\mathbb E[\psi\psi^\top]\succ0$, and the embedding differences $\phi(y)-\phi(z)$, $y,z\in V$, span $\mathbb R^d$. The second is exactly identifiability of $A_\theta$: if some $v\neq0$ had $\phi(y)^\top v$ constant in $y$, then $\Delta=vu^\top$ would leave every conditional unchanged for every $u\in\mathbb R^m$. Together they give $\Sigma(\eta)\succ0$ for every $\eta$ and hence $F\succ0$, $H_T\succ0$: for $\Delta\neq0$, $\langle\Delta,F\Delta\rangle=\mathbb E[(\Delta\psi)^\top\Sigma(A_{\theta^*_{\mathrm{post}}}\psi)(\Delta\psi)]>0$ because $G\succ0$.

\paragraph{The non-degeneracy condition of Theorem~\ref{thm:loglinear-bias-variance}.}
Assume in addition that $B\neq0$ and that (A2) $D:=\partial_TH_T|_{T=1}$ is not a multiple of $F$. The claim about $T=1$ additionally excludes a closed Lebesgue-null cone of teacher gaps $B$, which cannot be dispensed with: Remark~\ref{rem:t1-can-be-optimal} exhibits a model in which $T=1$ is the unique optimizer within the family.
We will show that under these conditions the inequality in \eqref{eq:loglinear-interior} is strict: some finite $(\lambda,T)$ with $T\neq1$ makes \methodname{} strictly better than maximum likelihood, than the frozen teacher and than every knowledge-distillation run. We do not claim that the infimum is attained, and Remark~\ref{rem:no-finite-optimum} shows that it need not be: the reachable penalty shapes form a curve that is open at its hot end.

Throughout, $\Phi\in\mathbb R^{|V|\times d}$ stacks the token embeddings $\phi(z)^\top$, so that $\Lambda(\eta)=\operatorname{lse}(\Phi\eta)$ with $\operatorname{lse}(u)=\log\sum_z e^{u_z}$, $\nabla\Lambda(\eta)=\Phi^\top\sigma(\Phi\eta)=\mu(\eta)$ and $\nabla^2\Lambda(\eta)=\Phi^\top\Sigma(\sigma(\Phi\eta))\Phi=\Sigma(\eta)$, where $\sigma$ is the softmax and $\Sigma(p)=\operatorname{diag}(p)-pp^\top$. We identify $A_\theta\in\mathbb R^{d\times m}$ with $\operatorname{vec}(A_\theta)$, so that for a linear map $\theta\mapsto A_\theta\psi$ the chain rule gives $\nabla_\theta h(A_\theta\psi)=\nabla h(A_\theta\psi)\,\psi^\top$ and $\nabla_\theta^2 h(A_\theta\psi)=\nabla^2h(A_\theta\psi)\otimes\psi\psi^\top$.

\begin{proposition}[Structure of the objective]
\label{prop:loglinear-structure}
For every $\lambda>0$ and $T>0$:
\begin{enumerate}
\item[(i)] $\widehat J_{\lambda,T}$ is convex in $\theta$, and strictly convex along every direction $\Delta$ that changes some logit difference on some observed context, i.e., $\phi(y)^\top\Delta\psi_t\neq\phi(z)^\top\Delta\psi_t$ for some $t$ and $y,z\in V$. Hence its minimizer is unique up to directions that leave every conditional $p_\theta(\cdot\mid\vx_{<t})$ unchanged.
\item[(ii)] The temperature matching term is a Bregman divergence centered at the pre-trained model:
$\widehat R_T(\theta)=\frac1n\sum_t D_T(A_\theta;A_{\theta_{\mathrm{pre}}^*},\psi_t)+\mathrm{const}$, where
\[
D_T(A_\theta;A_{\theta_{\mathrm{pre}}^*},\psi)=\Lambda\!\big(\tfrac{A_\theta\psi}{T}\big)-\Lambda\!\big(\tfrac{A_{\theta_{\mathrm{pre}}^*}\psi}{T}\big)-\mu\!\big(\tfrac{A_{\theta_{\mathrm{pre}}^*}\psi}{T}\big)^{\!\top}\tfrac{(A_\theta-A_{\theta_{\mathrm{pre}}^*})\psi}{T}\;\ge 0,
\]
with equality iff $p_{T\theta}(\cdot\mid\vx_{<i})=p_{TA_{\theta_{\mathrm{pre}}^*}}(\cdot\mid\vx_{<i})$. Thus $\widehat J_{\lambda,T}$ is a penalized maximum likelihood with a data-dependent penalty that vanishes exactly at $A_{\theta_{\mathrm{pre}}^*}$.
\item[(iii)] (Moment matching.) Its gradient is
\[
\nabla_\theta\widehat J_{\lambda,T}(\theta)=\frac1n\sum_{t=1}^n\Big[\big(\mu(A_\theta\psi_t)-\phi(x_t)\big)+\frac\lambda T\big(\mu(\tfrac{A_\theta\psi_t}{T})-\mu(\tfrac{A_{\theta_{\mathrm{pre}}^*}\psi_t}{T})\big)\Big]\psi_t^\top ,
\]
so at a minimizer the model's mean embedding matches the data on average, corrected by $\lambda/T$ times the gap between the student's and teacher's tempered mean embeddings.
\item[(iv)] (Curvature.) Its Hessian, as an operator on $\mathbb R^{d\times m}$, is
\[
\nabla^2_\theta\widehat J_{\lambda,T}(\theta)=\frac1n\sum_{t=1}^n\Big[\Sigma(A_\theta\psi_t)+\frac{\lambda}{T^2}\Sigma\!\big(\tfrac{A_\theta\psi_t}{T}\big)\Big]\otimes\psi_t\psi_t^\top ,
\]
and as $T\to\infty$, $\Sigma(A_\theta\psi_t/T)\to\Sigma_0:=\operatorname{Cov}_{\mathrm{unif}(V)}[\phi(y)]$ for every $\theta$. The teacher therefore contributes curvature $\tfrac{\lambda}{T^2}\,\Sigma_0\otimes\widehat G+o(T^{-2})$ with $\widehat G=\frac1n\sum_t\psi_t\psi_t^\top$: a ridge penalty toward $A_{\theta_{\mathrm{pre}}^*}$ whose strength is set by $\lambda/T^2$ and whose geometry is fixed by the embeddings and the context covariance.
\end{enumerate}
\end{proposition}

\begin{proof}[Proof of Proposition~\ref{prop:loglinear-structure}]
\emph{(i)} $\operatorname{lse}$ is convex, and $u\mapsto\Phi A_\theta\psi_t$ and $u\mapsto\Phi A_\theta\psi_t/T$ are linear in $\theta$, so every summand of \eqref{eq:loglinear-objective} is convex in $\theta$; the linear terms do not affect convexity. For strictness, $\nabla^2\operatorname{lse}(u)=\Sigma(\sigma(u))$ and $v^\top\Sigma(\sigma(u))v=\operatorname{Var}_{\sigma(u)}(v)$, which is positive unless $v$ is constant on $V$. Along a direction $\Delta$, the second directional derivative of the $t$-th clean summand is $\operatorname{Var}_{p_\theta(\cdot\mid\vx_{<t})}\!\big(\phi(y)^\top\Delta\psi_t\big)$, positive iff $y\mapsto\phi(y)^\top\Delta\psi_t$ is non-constant; the matching summand contributes the same quantity at temperature $T$ times $\lambda/T^2$. Summing over $t$ gives strict convexity along every $\Delta$ that changes a logit difference on some observed context, and flatness along the others; the latter are exactly the directions that leave every conditional unchanged, since a conditional depends on $A_\theta\psi_t$ only through its logit differences.

\emph{(ii)} Expanding $D_T$ and comparing with the $t$-th matching summand,
$\Lambda(\tfrac{A_\theta\psi_t}{T})-\mu(\tfrac{A_{\theta_{\mathrm{pre}}^*}\psi_t}{T})^\top\tfrac{A_\theta\psi_t}{T}
=D_T(A_\theta;A_{\theta_{\mathrm{pre}}^*},\psi_t)+\Lambda(\tfrac{A_{\theta_{\mathrm{pre}}^*}\psi_t}{T})-\mu(\tfrac{A_{\theta_{\mathrm{pre}}^*}\psi_t}{T})^\top\tfrac{A_{\theta_{\mathrm{pre}}^*}\psi_t}{T}$,
and the last two terms do not depend on $\theta$. $D_T$ is the Bregman divergence of the convex function $\eta\mapsto\Lambda(\eta/T)$ between $A_\theta\psi_t$ and $A_{\theta_{\mathrm{pre}}^*}\psi_t$, hence nonnegative, and zero iff $\Lambda(\cdot/T)$ is affine on the segment joining them, i.e., iff $\Phi(A_\theta-A_{\theta_{\mathrm{pre}}^*})\psi_t$ is constant on $V$ by the variance computation in (i), i.e., iff the two tempered conditionals coincide. Equivalently $D_T(A_\theta;A_{\theta_{\mathrm{pre}}^*},\psi_t)=D_{\mathrm{KL}}\big(p_{TA_{\theta_{\mathrm{pre}}^*}}(\cdot\mid\vx_{<t})\,\|\,p_{T\theta(\cdot\mid\vx_{<t})}\big)$, the standard identity between Bregman divergences of the log-partition and KL divergences in an exponential family.

\emph{(iii)} and \emph{(iv)} follow from the chain rule stated above applied to $\Lambda(A_\theta\psi_t)$, $\phi(x_t)^\top A_\theta\psi_t$, $\Lambda(A_\theta\psi_t/T)$ and $\mu(\cdot)^\top A_\theta\psi_t/T$; the factor $1/T$ from differentiating $A_\theta\psi_t/T$ appears once in the gradient and twice in the Hessian. Finally $\sigma(\Phi\eta/T)\to\sigma(0)=\mathbf 1/|V|$ as $T\to\infty$ for every fixed $\eta$, so $\Sigma(\eta/T)\to\Phi^\top\Sigma(\mathbf 1/|V|)\Phi=\operatorname{Cov}_{\mathrm{unif}(V)}[\phi(y)]=\Sigma_0$, uniformly on bounded sets of $\eta$.
\end{proof}

\noindent \textbf{Setting for Theorem~\ref{thm:loglinear-bias-variance}.}
Let $\ell_t(\theta)$ denote the $t$-th summand of \eqref{eq:loglinear-objective} (clean plus $\lambda$ times matching), so $\widehat J_{\lambda,T}=\frac1n\sum_t\ell_t$ and $J_{\lambda,T}=\mathbb E[\ell_t]$, the expectation over $\psi\sim\mathcal D_x$ and $x\sim p_{A_{\theta^*_{\mathrm{post}}}}(\cdot\mid\psi)$. We consider the local regime $B=B_0/\sqrt n$ with $B_0$ fixed, so $\|B\|=O(n^{-1/2})$; all $O(\cdot)$ constants below are uniform.
Write $H_T^{\mathrm{pre}}=\nabla^2R_T(\theta_{\mathrm{pre}}^*)=\frac1{T^2}\mathbb E[\Sigma(A_{\theta_{\mathrm{pre}}^*}\psi/T)\otimes\psi\psi^\top]$ for the curvature of the penalty at its center; $H_T^{\mathrm{pre}}=H_T+O(\|B\|)$ by smoothness of $\Sigma$. $F$ and $H_T$ are positive definite by the non-degeneracy conditions above.

\begin{lemma}[Consistency, asymptotic normality, second moments]
\label{lem:loglinear-clt}
For fixed $\lambda$, $T$ and $B$: $\widehat\theta_\lambda\to\theta^*_\lambda$ almost surely and $\sqrt n(\widehat\theta_\lambda-\theta^*_\lambda)\Rightarrow\mathcal N(0,F_\lambda^{-1}V_\lambda F_\lambda^{-1})$, with $F_\lambda=\nabla^2J_{\lambda,T}(\theta^*_\lambda)$ and $V_\lambda=\operatorname{Cov}[\nabla\ell_t(\theta^*_\lambda)]$. Moreover $V_\lambda=F+O(\|B\|)$, $F_\lambda=F+\lambda H_T+O(\|B\|)$, $\mathbb E[\widehat\theta_\lambda]-\theta^*_\lambda=O(1/n)$ and $n\,\mathbb E\|\widehat\theta_\lambda-\theta^*_\lambda\|^2=\operatorname{tr}(F_\lambda^{-1}V_\lambda F_\lambda^{-1})+O(n^{-1/2})$.
\end{lemma}
\begin{proof}
Each $\ell_t$ is convex and $C^\infty$ in $\theta$ with bounded derivatives on bounded sets (the logits are bounded by $\|A_\theta\|\,\|\phi\|\,\|\psi\|$), and $J_{\lambda,T}$ is strictly convex with positive definite Hessian $\nabla^2J_{\lambda,T}(\theta)=\mathbb E[\Sigma(A_\theta\psi)\otimes\psi\psi^\top]+\frac{\lambda}{T^2}\mathbb E[\Sigma(\tfrac{A_\theta\psi}{T})\otimes\psi\psi^\top]$, so it has a unique minimizer $\theta^*_\lambda$. Pointwise convergence $\widehat J_{\lambda,T}\to J_{\lambda,T}$ (strong law) together with convexity gives uniform convergence on compacts and hence $\widehat\theta_\lambda\to \theta^*_\lambda$ almost surely. Asymptotic normality is the classical result for smooth M-estimators: expanding the first-order condition $\sum_t\nabla\ell_t(\widehat\theta_\lambda)=0$ around $\theta^*_\lambda$,
$0=\frac1{\sqrt n}\sum_t\nabla\ell_t(\theta^*_\lambda)+\Big[\frac1n\sum_t\nabla^2\ell_t(\theta^*_\lambda)+o_P(1)\Big]\sqrt n(\widehat\theta_\lambda-\theta^*_\lambda)$,
where the first term is a sum of i.i.d. mean-zero (by optimality of $\theta^*_\lambda$) bounded vectors and converges to $\mathcal N(0,V_\lambda)$ with $V_\lambda=\operatorname{Cov}[\nabla\ell_t(\theta^*_\lambda)]$, and the bracket converges to $F_\lambda=\nabla^2J_{\lambda,T}(\theta^*_\lambda)$. Slutsky gives $\sqrt n(\widehat\theta_\lambda-\theta^*_\lambda)\Rightarrow\mathcal N(0,F_\lambda^{-1}V_\lambda F_\lambda^{-1})$. Writing $\nabla\ell_t(\theta)=(\mu(A_\theta\psi_t)-\phi(x_t))\psi_t^\top+\frac\lambda T(\mu(\tfrac{A_\theta\psi_t}{T})-\mu(\tfrac{A_{\theta_{\mathrm{pre}}^*}\psi_t}{T}))\psi_t^\top$ split it at $\theta=\theta^*_{\mathrm{post}}$ into the clean part $s_{\mathrm{clean}}=(\mu(A_{\theta^*_{\mathrm{post}}}\psi_t)-\phi(x_t))\psi_t^\top$ and the teacher part $r_T(\psi_t)=\frac\lambda T(\mu(\tfrac{A_{\theta^*_{\mathrm{post}}}\psi_t}{T})-\mu(\tfrac{A_{\theta^*_{\mathrm{pre}}}\psi_t}{T}))\psi_t^\top$, which is a deterministic function of $\psi_t$. Conditionally on $\psi_t$ the only randomness is $\phi(x_t)$, with covariance $\Sigma(A_{\theta^*_{\mathrm{post}}}\psi_t)$, so $s_{\mathrm{clean}}$ has conditional mean zero and the law of total covariance gives
$\operatorname{Cov}[\nabla\ell_t(\theta^*_{\mathrm{post}})]=\mathbb E[\Sigma(A_{\theta^*_{\mathrm{post}}}\psi_t)\otimes\psi_t\psi_t^\top]+\operatorname{Cov}(r_T(\psi_t))=F+O(\|B\|^2)$,
the second term being $O(\|B\|^2)$ because $r_T=O(\|B\|)$ uniformly. Moving the evaluation point from $\theta^*_{\mathrm{post}}$ to $\theta^*_\lambda$, and since $A_{\theta^*_\lambda}-A_{\theta^*_{\mathrm{post}}}=O(\|B\|)$ by Lemma~\ref{lem:loglinear-bias} and $\nabla\ell_t$ is smooth, $V_\lambda=F+O(\|B\|)$ and $F_\lambda=F+\lambda H_T+O(\|B\|)$.

For the second moments, one cannot simply condition on the prefixes: a finite sample may miss feature directions, and then $\widehat J_{\lambda,T}$ is not strongly convex for that realization. Work instead on the event $E_n=\{\widehat G_n\succeq\tfrac12G\}$, where $\widehat G_n=\frac1n\sum_t\psi_t\psi_t^\top$. The summands $\psi_t\psi_t^\top$ are i.i.d.\ and positive semidefinite with $\|\psi_t\psi_t^\top\|_{\mathrm{op}}\le R_\psi^2$ and mean $G\succ0$, so the matrix Chernoff bound gives $\mathbb P(E_n^c)\le m\,e^{-cn}$ with $c=\lambda_{\min}(G)/(8R_\psi^2)$. On $E_n$ the empirical objective is strongly convex on $\mathcal B_R$: the set $\{A_\theta\psi:\theta\in\mathcal B_R,\|\psi\|\le R_\psi\}$ is compact and $\Sigma(\cdot)\succ0$ on it by the two non-degeneracy conditions, so $\sigma_0:=\inf\lambda_{\min}(\Sigma(A_\theta\psi))>0$ there and, for every $\Delta$,
$\langle\Delta,\nabla^2\widehat J_{\lambda,T}(\theta)\Delta\rangle\ge\frac1n\sum_t(\Delta\psi_t)^\top\Sigma(A_\theta\psi_t)(\Delta\psi_t)\ge\sigma_0\operatorname{tr}(\Delta\widehat G_n\Delta^\top)\ge\tfrac12\sigma_0\lambda_{\min}(G)\|\Delta\|_F^2$, the first inequality discarding the matching term of the Hessian, which is positive semidefinite.
With that modulus $\kappa$, the two-point inequality at the interior minimizer $\theta^*_\lambda$ of $J_{\lambda,T}$ gives $\kappa\|\widehat\theta_\lambda-\theta^*_\lambda\|\le\|\nabla\widehat J_{\lambda,T}(\theta^*_\lambda)\|$ on $E_n$, and $\nabla\widehat J_{\lambda,T}(\theta^*_\lambda)$ is an average of i.i.d.\ bounded terms with mean $\nabla J_{\lambda,T}(\theta^*_\lambda)=0$, so $\mathbb E\|\nabla\widehat J_{\lambda,T}(\theta^*_\lambda)\|^2=O(1/n)$ and $\mathbb E[\|\widehat\theta_\lambda-\theta^*_\lambda\|^2\mathbf 1_{E_n}]\le C/n$. On $E_n^c$ we use only that both iterates lie in $\mathcal B_R$, an operator-norm ball, so $\|A_{\widehat\theta_\lambda}-A_{\theta^*_\lambda}\|_F\le2R\sqrt{\min(d,m)}$ and that part contributes at most $4R^2\min(d,m)\,m\,e^{-cn}$, which is $o(n^{-p})$ for every $p$; hence $\mathbb E\|\widehat\theta_\lambda-\theta^*_\lambda\|^2\le C'/n$ for every $n$. The same event carries the refinement that follows: all derivatives of $\ell_t$ are bounded on $\mathcal B_R$, so the one-step expansion of the first-order condition can be carried one order further on $E_n$: $\widehat\theta_\lambda-\theta^*_\lambda=-F_\lambda^{-1}\frac1n\sum_t\nabla\ell_t(\theta^*_\lambda)+\Delta_n$ with $\mathbb E[\|\Delta_n\|^2\mathbf 1_{E_n}]\le C/n^2$ and the complement again exponentially negligible. That bound is where the fourth moments enter: writing $S_n=\frac1n\sum_t\nabla\ell_t(\theta^*_\lambda)$ for the empirical score and $Q_n=\frac1n\sum_t\nabla^2\ell_t(\theta^*_\lambda)-F_\lambda$ for the Hessian fluctuation, both are averages of bounded i.i.d.\ mean-zero terms, so $\mathbb E\|S_n\|^4=O(n^{-2})$ and $\mathbb E\|Q_n\|^4=O(n^{-2})$; on $E_n$ the curvature bound gives $\|\widehat\theta_\lambda-\theta^*_\lambda\|\le\kappa^{-1}\|S_n\|$, and the second-order Taylor remainder of the first-order condition is bounded by $\|Q_n\|\|\widehat\theta_\lambda-\theta^*_\lambda\|$ plus a third-derivative term of order $\|\widehat\theta_\lambda-\theta^*_\lambda\|^2$, so $\mathbb E[\|\Delta_n\|^2\mathbf 1_{E_n}]=O(\mathbb E\|Q_n\|^4)^{1/2}O(\mathbb E\|S_n\|^4)^{1/2}+O(\mathbb E\|S_n\|^4)=O(n^{-2})$ by Cauchy--Schwarz. Hence $\mathbb E\|\widehat\theta_\lambda-\theta^*_\lambda\|^2=\frac1n\operatorname{tr}(F_\lambda^{-1}V_\lambda F_\lambda^{-1})+O(n^{-3/2})$ by Cauchy--Schwarz on the cross term, and $\mathbb E[\widehat\theta_\lambda]-\theta^*_\lambda=\mathbb E[\Delta_n]=O(1/n)$.
\end{proof}

\begin{lemma}[Population bias]
\label{lem:loglinear-bias}
$b_\lambda:=A_{\theta^*_\lambda}-A_{\theta^*_{\mathrm{post}}}=-\lambda(F+\lambda H_T^{\mathrm{pre}})^{-1}H_T^{\mathrm{pre}}B+O(\|B\|^2)$ as $B\to0$, and $b_0=0$.
\end{lemma}
\begin{proof}
Let $L(\theta)=\mathbb E[\Lambda(A_\theta\psi)-\phi(x)^\top A_\theta\psi]$ and $R_T(\theta)=\mathbb E[D_T(A_\theta;A_{\theta_{\mathrm{pre}}^*},\psi)]$, so $J_{\lambda,T}=L+\lambda R_T+\mathrm{const}$. Since $\mathbb E[\phi(x)\mid\psi]=\mu(A_{\theta^*_{\mathrm{post}}}\psi)$, $\nabla L(\theta)=\mathbb E[(\mu(A_\theta\psi)-\mu(A_{\theta^*_{\mathrm{post}}}\psi))\psi^\top]$, which vanishes at $A_{\theta^*_{\mathrm{post}}}$ with $\nabla^2L(A_{\theta^*_{\mathrm{post}}})=F$; likewise $\nabla R_T(\theta)=\frac1T\mathbb E[(\mu(\tfrac{A_\theta\psi}{T})-\mu(\tfrac{A_{\theta_{\mathrm{pre}}^*}\psi}{T}))\psi^\top]$ vanishes at $A_{\theta_{\mathrm{pre}}^*}$ with $\nabla^2R_T(A_{\theta_{\mathrm{pre}}^*})=H_T^{\mathrm{pre}}$. The minimizer solves $\nabla L(\theta^*_\lambda)+\lambda\nabla R_T(\theta^*_\lambda)=0$. Taylor-expanding $\nabla L$ around $A_{\theta^*_{\mathrm{post}}}$ and $\nabla R_T$ around $A_{\theta_{\mathrm{pre}}^*}$, with $b_\lambda=A_{\theta^*_\lambda}-A_{\theta^*_{\mathrm{post}}}$ and $A_{\theta^*_\lambda}-A_{\theta_{\mathrm{pre}}^*}=b_\lambda+B$,
$F b_\lambda+\lambda H_T^{\mathrm{pre}}(b_\lambda+B)+O(\|b_\lambda\|^2+\|b_\lambda+B\|^2)=0$.
For $B=0$ the unique solution is $b_0=0$ (then $A_{\theta^*_{\mathrm{post}}}=A_{\theta_{\mathrm{pre}}^*}$ minimizes both terms). For small $B$, the implicit function theorem applied to the smooth map $(\theta,B)\mapsto\nabla L(\theta)+\lambda\nabla R_T(\theta)$, whose $\theta$-derivative $F+\lambda H_T^{\mathrm{pre}}$ is invertible, gives $b_\lambda=-\lambda(F+\lambda H_T^{\mathrm{pre}})^{-1}H_T^{\mathrm{pre}}B+O(\|B\|^2)$.
\end{proof}

\begin{proof}[Proof of Theorem~\ref{thm:loglinear-bias-variance}]
\emph{The expansion.} Write $R_n(\lambda,T)=\mathbb E\|A_{\widehat\theta_{\lambda,T}}-A_{\theta^*_{\mathrm{post}}}\|_F^2$ for the risk. Decompose $A_{\widehat\theta_\lambda}-A_{\theta^*_{\mathrm{post}}}=(\widehat\theta_\lambda-\theta^*_\lambda)+b_\lambda$, so $R_n(\lambda)=\mathbb E\|\widehat\theta_\lambda-\theta^*_\lambda\|^2+\|b_\lambda\|^2+2\langle\mathbb E[\widehat\theta_\lambda]-\theta^*_\lambda,b_\lambda\rangle$. By Lemma~\ref{lem:loglinear-clt} and $\|B\|=O(n^{-1/2})$, the first term is $\frac1n\operatorname{tr}\big((F+\lambda H_T)^{-1}F(F+\lambda H_T)^{-1}\big)+O(n^{-3/2})$. By Lemma~\ref{lem:loglinear-bias} with $B=B_0/\sqrt n$ and $H_T^{\mathrm{pre}}=H_T+O(n^{-1/2})$, the second is $\frac{\lambda^2}n\|(F+\lambda H_T)^{-1}H_TB_0\|_F^2+O(n^{-3/2})$. The cross term is $O(1/n)\cdot O(n^{-1/2})$. Since $B_0=\sqrt n\,B$, this is exactly \eqref{eq:loglinear-risk}; uniformity in $\lambda$ on compacts holds because every constant above is continuous in $\lambda$. For the rest of the proof $\rho$ denotes the leading term of \eqref{eq:loglinear-risk} itself, and $\bar\rho=n\rho$ its rescaling, in which the maximum-likelihood value is $\operatorname{tr}(F^{-1})$ and the teacher gap enters as $B_0=\sqrt n\,B$. The strictness argument below is stated for $\rho$, the window calculation and the aligned Example for $\bar\rho$.

\emph{Endpoints.} In the rescaled units, $\bar\rho(0)=\operatorname{tr}(F^{-1})$ is immediate; unscaled this reads $\rho(0,T)=\operatorname{tr}(F^{-1})/n$. As $\lambda\to\infty$, $(F+\lambda H_T)^{-1}\lambda H_T=I-(F+\lambda H_T)^{-1}F\to I$ because $\|(F+\lambda H_T)^{-1}\|_{\mathrm{op}}\le1/(\lambda\lambda_{\min}(H_T))$, so the bias term tends to $\|B_0\|_F^2$, i.e.\ $\lim_{\lambda\to\infty}\rho(\lambda,T)=\|B\|_F^2$, and the variance is at most $\|F\|_{\mathrm{op}}\operatorname{tr}\big((F+\lambda H_T)^{-2}\big)\to0$.

\emph{Derivative at zero (Remark~\ref{rem:loglinear-window}).} The bias term is $O(\lambda^2)$ and contributes nothing. With $M_\lambda=(F+\lambda H_T)^{-1}$ we have $M_0'=-F^{-1}H_TF^{-1}$, so $\frac{d}{d\lambda}\operatorname{tr}(M_\lambda FM_\lambda)\big|_0=2\operatorname{tr}(M_0'FM_0)=-2\operatorname{tr}(F^{-1}H_TF^{-1}FF^{-1})=-2\operatorname{tr}(F^{-2}H_T)$, which is negative because $F^{-2}\succ0$ and $H_T\succ0$; equivalently $\partial_\lambda\bar\rho(0,T)=-2\operatorname{tr}(F^{-2}H_T)$ and $\partial_\lambda\rho(0,T)=-2\operatorname{tr}(F^{-2}H_T)/n$.

\emph{Strictness.} In two steps: the $T=1$ slice already beats the other two extremes, and moving $T$ off $1$ beats the $T=1$ slice for generic $B$.

\emph{(i) The $T=1$ slice.} Setting $T=1$ in the definition of $H_T$ gives $H_1=\mathbb E[\Sigma(A_{\theta^*_{\mathrm{post}}}\psi)\otimes\psi\psi^\top]=F$. Hence $M=\lambda F$, the shrinkage operator is $(F+M)^{-1}M=\alpha I$ with $\alpha=\lambda/(1+\lambda)$, and \eqref{eq:loglinear-risk} becomes, for every $F$ and $B$,
\[
\rho(\lambda,1)=(1-\alpha)^2V+\alpha^2W,\qquad V=\frac{\operatorname{tr}(F^{-1})}{n},\quad W=\|B\|_F^2 ,
\]
a scalar bias--variance tradeoff. It is strictly convex in $\alpha$ and minimized at $\alpha^\circ=V/(V+W)\in(0,1)$, i.e.\ at the interior point $\lambda^*_1=V/W\in(0,\infty)$, where $\partial_\lambda\rho(\lambda^*_1,1)=0$. Its value there is the harmonic-type mean $VW/(V+W)$, which is strictly smaller than both $V$ and $W$ whenever $V,W>0$, that is, whenever $B\neq0$. Since $V$ is the maximum-likelihood risk $\rho(0,T)$ and $W=\lim_{\lambda\to\infty}\rho(\lambda,T)$ is the frozen-teacher risk, the best $T=1$ run already beats those two strictly. No further assumption is needed for this step.

\emph{(ii) Off $T=1$.} Regard $\rho$ as a function of the penalty $M\succeq0$ alone. For a symmetric direction $\Delta$, differentiating $(F+M)^{-1}M=I-(F+M)^{-1}F$ and $(F+M)^{-1}F(F+M)^{-1}$ at $M=\lambda F$, where $(F+M)^{-1}=F^{-1}/(1+\lambda)$, gives
\begin{equation}
\big\langle\nabla_M\rho(\lambda F;B),\Delta\big\rangle=\frac{2}{(1+\lambda)^3}\Big[\lambda\,\langle B,F^{-1}\Delta B\rangle-\frac1n\operatorname{tr}(F^{-2}\Delta)\Big].
\label{eq:gradM-at-lambdaF}
\end{equation}
Taking $\Delta=F$ returns $2(\lambda W-V)/(1+\lambda)^3$, which vanishes exactly at $\lambda^*_1=V/W$, as it must. Take instead $\Delta=D$. Since $\partial_\lambda M=H_1=F$ and $\partial_TM|_{T=1}=\lambda^*_1D$, and since $\partial_\lambda\rho(\lambda^*_1,1)=0$, we get $\partial_T\rho(\lambda^*_1,1)=\lambda^*_1\Pi(B)$ with $\Pi(B)=\langle\nabla_M\rho(\lambda^*_1(B)F;B),D\rangle$. Writing $b=\operatorname{vec}(B)$, $r=b^\top b=\|B\|_F^2$ and $s=\operatorname{tr}(F^{-1})$, so that $\lambda^*_1=s/(nr)$, equation \eqref{eq:gradM-at-lambdaF} collects into a quadratic form in $b$:
\begin{equation}
\Pi(B)=\frac{2\,b^\top Cb}{n\,r\,(1+\lambda^*_1)^3},\qquad C=s\cdot\operatorname{sym}(F^{-1}D)-\operatorname{tr}(F^{-2}D)\,I ,
\label{eq:pi-quadratic}
\end{equation}
using $b^\top F^{-1}Db=b^\top\operatorname{sym}(F^{-1}D)b$. The matrix $C$ depends on $F$ and $D$ only, and $C=0$ if and only if $D$ is a multiple of $F$: in an eigenbasis of $F$ with eigenvalues $\mu_i$ one has $(\operatorname{sym}(F^{-1}D))_{ij}=\tfrac12D_{ij}(\mu_i^{-1}+\mu_j^{-1})$, so $C=0$ forces $D_{ij}=0$ for $i\neq j$ and $D_{ii}=\mu_i\operatorname{tr}(F^{-2}D)/s$, i.e.\ $D=cF$; conversely $D=cF$ gives $\operatorname{sym}(F^{-1}D)=cI$ and $\operatorname{tr}(F^{-2}D)=cs$, hence $C=0$. Thus (A2) says precisely $C\neq0$, and then $Z=\{B:b^\top Cb=0\}$ is the zero set of a nonzero quadratic form: closed, of Lebesgue measure zero, and a cone, so membership of $Z$ is unaffected by the $n^{-1/2}$ scaling of $B$. For $B\notin Z$ we have $\partial_T\rho(\lambda^*_1,1)\neq0$, so moving $T$ off $1$ in the descent direction strictly decreases $\rho$ below $\min_\lambda\rho(\lambda,1)$, and by step (i) below all three extremes. The descent direction is $T>1$ when $b^\top Cb<0$ and $T<1$ when $b^\top Cb>0$; the theorem claims improvement, not that it is always on the hot side. Note that this argument needs no commutativity between $F$, $H_T$ and $D$, and no analyticity: $b^\top Cb$ is a polynomial.

\emph{The weak statement \eqref{eq:loglinear-interior}.} It is pure continuity, and needs neither $B\neq0$ nor (A2). $\rho(\cdot,T)$ is continuous on $[0,\infty)$ and has a finite limit at $\infty$, so $\inf_{\lambda\in(0,\infty)}\rho(\lambda,T)\le\min\{\rho(0,T),\lim_{\lambda\to\infty}\rho(\lambda,T)\}$, both extremes being limits of interior values; the same argument in $T$ gives $\inf_{T\neq1}\rho(\lambda,T)\le\rho(\lambda,1)$ and $\inf_{T<\infty}\rho(\lambda,T)\le\lim_{T\to\infty}\rho(\lambda,T)$. Taking the smallest of the three right-hand sides gives \eqref{eq:loglinear-interior}; the assumptions are what turn it into a strict inequality.

\emph{The window $\lambda_0$ (Remark~\ref{rem:loglinear-window}).} Write $F+\lambda H_T=F^{1/2}(I+\lambda K)F^{1/2}$ with $K=F^{-1/2}H_TF^{-1/2}\succ0$ and eigenpairs $(k_i,u_i)$. Then $M_\lambda FM_\lambda=F^{-1/2}(I+\lambda K)^{-2}F^{-1/2}$, so the variance equals $\sum_i(1+\lambda k_i)^{-2}\,u_i^\top F^{-1}u_i$. Since $(1-2x+3x^2)(1+x)^2-1=4x^3+3x^4\ge0$, we have $(1+x)^{-2}\le1-2x+3x^2$ for $x\ge0$, whence the variance is at most $\operatorname{tr}(F^{-1})-2\lambda\operatorname{tr}(KF^{-1})+3\lambda^2\operatorname{tr}(K^2F^{-1})=\operatorname{tr}(F^{-1})-2\lambda\operatorname{tr}(H_TF^{-2})+3\lambda^2\operatorname{tr}(H_TF^{-1}H_TF^{-2})$. The bias term is at most $\lambda^2\|M_\lambda\|_{\mathrm{op}}^2\|H_TB_0\|_F^2\le\lambda^2\|H_TB_0\|_F^2/\lambda_{\min}(F)^2$. Hence $\bar\rho(\lambda)-\bar\rho(0)\le-2\lambda\operatorname{tr}(H_TF^{-2})+\lambda^2\big(\|H_TB_0\|_F^2/\lambda_{\min}(F)^2+3\operatorname{tr}(H_TF^{-1}H_TF^{-2})\big)$, which is negative for $0<\lambda<\lambda_0$. Finally, for fixed $\lambda\in(0,\lambda_0)$, $n\big(R_n(\lambda)-R_n(0)\big)=\bar\rho(\lambda)-\bar\rho(0)+O(n^{-1/2})<0$ once $n$ is large enough.
\end{proof}

\begin{remark}[How much teacher, explicitly]
\label{rem:loglinear-window}
The improvement over maximum likelihood comes from $\partial_\lambda\rho(0,T)=-2\operatorname{tr}(F^{-2}H_T)/n<0$, the bias entering only at second order in $\lambda$; at the other end, $\rho(\lambda,T)\to\|B\|_F^2$ as $\lambda\to\infty$, pure bias. Carrying both terms explicitly, every $\lambda$ below
\[
\lambda_0=\frac{2\operatorname{tr}(H_TF^{-2})}{n\|H_TB\|_F^2/\lambda_{\min}(F)^2+3\operatorname{tr}(H_TF^{-1}H_TF^{-2})},
\]
improves on maximum likelihood; this window is of order $\operatorname{tr}(F^{-2}H_T)/(n\|B\|^2)$ when the teacher gap dominates.
\end{remark}

\begin{remark}[$T=1$ can be optimal]
\label{rem:t1-can-be-optimal}
The exceptional set $Z$ is not empty. Take $V=\{0,1\}$ with scalar embeddings $\phi(0)=0$, $\phi(1)=1$, two equally likely context features $e_1,e_2$, and target logits $0$ and $2$, so $F$ and every $H_T$ are diagonal, with $\mu_1=\tfrac18$ and $\mu_2\approx0.0525$. Choose the teacher gap with $b_j^2=1/(n\mu_j)$, which is of the order $n^{-1/2}$ the theorem assumes. Each coordinate then contributes $\big((1-\alpha_j)^2+\alpha_j^2\big)/(n\mu_j)$ to $\rho$, minimized at $\alpha_j=\tfrac12$; since $H_1=F$, the single choice $\lambda=1$ attains that in both coordinates at once, so $\lambda H_1=F$ minimizes $\rho$ over all \emph{diagonal} penalties, and in particular over the family $\{\lambda H_T\}$, whose every member is diagonal here. Moreover $\kappa_1(T)/\kappa_2(T)$ is strictly decreasing in $T$, so $\lambda H_T=F$ has the unique solution $(\lambda,T)=(1,1)$: knowledge distillation is not merely tied here but is the unique optimizer. (A2) holds, the diagonal ratios of $D$ to $F$ being $-2$ and $\approx-0.477$; what fails is $B\notin Z$. Indeed $n\mu_jb_j^2$ is constant across directions, which makes $\lambda^*_1=V/W=1$ and, by \eqref{eq:pi-quadratic}, $b^\top Cb=\tfrac1n\big[(\sum_k\mu_k^{-1})(\sum_j d_j\mu_j^{-2})-(\sum_j\mu_j^{-1})(\sum_k d_k\mu_k^{-2})\big]=0$, so $\Pi(B)=0$ and $T=1$ is stationary. Note that $\Pi(B)=0$ is weaker than $\nabla_M\rho(\lambda^*_1F)=0$, which fails here: by \eqref{eq:gradM-at-lambdaF} the gradient has the off-diagonal entries of $\operatorname{sym}(F^{-1}bb^\top)$, which are nonzero, so $F$ is optimal within the reachable family but not among all positive-definite penalties. The coincidence $b^\top Cb=0$ has measure zero in $B$, which is what the genericity caveat excludes.
\end{remark}

\begin{remark}[The infimum need not be attained]
\label{rem:no-finite-optimum}
Theorem~\ref{thm:loglinear-bias-variance} asserts a strict inequality, not the existence of a best $(\lambda,T)$, and the distinction is real. The risk depends on the pair only through $M=\lambda H_T$, so $\lambda$ sets the scale of the penalty and $T$ its shape, and the reachable shapes trace the curve $T\mapsto T^2H_T$, which starts at $F$ at $T=1$ and tends to $\Sigma_0\otimes G$ as $T\to\infty$. Convergence by itself decides nothing about attainment: if the target conditionals are uniform then $T^2H_T=\Sigma_0\otimes G$ at every temperature and the curve is a single point. But the limiting shape need not be reachable at any finite $T$, and then the joint infimum can fail to be attained even under the non-degeneracy condition. The following example does that, on an open set of teacher gaps. Keep the model of Remark~\ref{rem:t1-can-be-optimal} but take $b_1^2=2/n$ and $b_2^2=1/n$. By the Example below the best penalty eigenvalues are $m_j^*=1/(nb_j^2)$, that is $(\tfrac12,1)$, whereas the reachable ratio is $m_1/m_2=\kappa_1(T)/\kappa_2(T)=\cosh^2(1/T)$, which exceeds $1$ at every finite $T$ and tends to $1$ only as $T\to\infty$. With $\lambda$ free to set the scale, the reachable set is therefore exactly the open cone $\{m_1>m_2>0\}$, and the unconstrained optimum $(\tfrac12,1)$ lies outside it. The optimum over the closure is a different point: each coordinate contributes $(m_j^2b_j^2+\mu_j/n)/(\mu_j+m_j)^2$, strictly decreasing then strictly increasing with minimum at $m_j^*$, so from any feasible $m_1>m_2$ one can move to the diagonal without increasing either contribution, and the constrained minimum sits on the face $m_1=m_2$. That face is disjoint from the reachable set, so the infimum over finite $(\lambda,T)$ is approached only as $T\to\infty$ with $\lambda\propto T^2$ and is never attained. Note that this is a different limit from the maximum-likelihood one in \eqref{eq:loglinear-interior}, which sends $T\to\infty$ at fixed $\lambda$ and does kill the penalty. Read positively, a missing minimizer says the model would like the hottest teacher available, with $\lambda$ raised to compensate for the flattening.
\end{remark}

\noindent \textbf{Example (aligned case).}
Suppose $F$ and $H_T$ share eigenvectors, with eigenvalues $(\mu_j)$ and $(\kappa_j)$, and let $b_j$ be the coordinates of $B_0$ in that basis. Then
\begin{equation}
\bar\rho(\lambda)=\sum_{j}\Big[(1-\alpha_j)^2\frac1{\mu_j}+\alpha_j^2b_j^2\Big],\qquad\alpha_j=\frac{\lambda\kappa_j}{\mu_j+\lambda\kappa_j}\in(0,1).
\label{eq:loglinear-bias-variance}
\end{equation}
Diagonalizing both terms of $\bar\rho$ in the common eigenbasis, direction $j$ contributes
$\big(\tfrac{\lambda\kappa_j}{\mu_j+\lambda\kappa_j}\big)^2b_j^2+\frac{\mu_j}{(\mu_j+\lambda\kappa_j)^2}=\alpha_j^2b_j^2+(1-\alpha_j)^2\frac1{\mu_j}$,
using $1-\alpha_j=\mu_j/(\mu_j+\lambda\kappa_j)$. Comparing with the maximum-likelihood value $1/\mu_j$: the shrunk direction is better iff $\alpha_j^2b_j^2<\big(1-(1-\alpha_j)^2\big)\frac1{\mu_j}=\alpha_j(2-\alpha_j)\frac1{\mu_j}$, i.e. iff $b_j^2<\frac{2-\alpha_j}{\alpha_j}\frac1{\mu_j}$. Minimizing $(1-\alpha)^2v+\alpha^2b^2$ over $\alpha$ with $v=1/\mu_j$ gives $\alpha^*=v/(v+b^2)=1/(1+\mu_jb_j^2)$. In the original units ($b_j=\sqrt n\,b_j'$ with $b_j'$ the coordinates of $B$) these read $b_j'^2<\frac{2-\alpha_j}{\alpha_j}\frac1{n\mu_j}$ and $\alpha_j^*=1/(1+n\mu_jb_j'^2)$: the amount of teacher to trust in a direction is set by the ratio of the teacher's squared error to the sampling variance $1/(n\mu_j)$ there. Differentiating at $\lambda=0$ recovers $\bar\rho'(0)=-2\sum_j\kappa_j/\mu_j^2$. Finally, since the shrinkage in direction $j$ depends on $(\lambda,T)$ only through the eigenvalue $m_j=\lambda\kappa_j$ of $M=\lambda H_T$, through $\alpha_j=m_j/(\mu_j+m_j)$, the optimal $\alpha_j^*$ corresponds to $m_j^*=\mu_j\alpha_j^*/(1-\alpha_j^*)=1/(n b_j'^2)$: the penalty should be strongest in the directions where the teacher's error $b_j'$ is smallest, a shape that depends on the teacher and not on the data, and one that $M=\lambda F$ (i.e.\ $T=1$) cannot produce unless $n\mu_jb_j'^2$ is constant across directions (Theorem~\ref{thm:loglinear-bias-variance}).

\begin{lemma}[Curvature of a softmax is at least its smallest probability]
\label{lem:softmax-curvature}
For every $\eta\in\mathbb R^d$, $\Sigma(\eta)\succeq e^{-\operatorname{range}(\Phi\eta)}\,\Sigma_0$, where $\operatorname{range}(u)=\max_zu_z-\min_zu_z$ and $\Sigma_0=\operatorname{Cov}_{\mathrm{unif}(V)}[\phi(y)]=\frac1{|V|}\Phi^\top(I-\tfrac1{|V|}\mathbf 1\mathbf 1^\top)\Phi$.
\end{lemma}
\begin{proof}
Let $p=\sigma(\Phi\eta)$ and $u\in\mathbb R^{|V|}$. Then $u^\top(\operatorname{diag}(p)-pp^\top)u=\operatorname{Var}_p(u)=\sum_zp_z(u_z-\bar u_p)^2\ge p_{\min}\sum_z(u_z-\bar u_p)^2\ge p_{\min}\sum_z(u_z-\bar u)^2=p_{\min}\,u^\top(I-\tfrac1{|V|}\mathbf 1\mathbf 1^\top)u$, where $\bar u_p$ is the $p$-weighted mean, $\bar u$ the unweighted mean, and the last inequality holds because the unweighted mean minimizes $c\mapsto\sum_z(u_z-c)^2$. Hence $\Sigma(\eta)=\Phi^\top(\operatorname{diag}(p)-pp^\top)\Phi\succeq p_{\min}|V|\,\Sigma_0$, and $p_{\min}=\min_ze^{u_z}/\sum_we^{u_w}\ge e^{-\operatorname{range}(u)}/|V|$ with $u=\Phi\eta$.
\end{proof}

\noindent \textbf{Dependence within documents.}
Theorem~\ref{thm:loglinear-bias-variance} is stated for independent prefixes for clarity, but nothing is lost when the $n=n_1l$ tokens come from $n_1$ independent documents of fixed length $l$, which is the situation of \eqref{eq:tm-objective}. Group the score by document, $Z_k=\frac1l\sum_{i=1}^l\nabla\ell_{k,i}(\theta^*_\lambda)$. The $Z_k$ are i.i.d.\ across $k$, so Lemma~\ref{lem:loglinear-clt} applies with $n_1\to\infty$ at fixed $l$, its empirical-curvature concentration now run on the document-level averages, and with constants that may depend on $l$. What has to be rechecked is the covariance. Per document it carries an extra factor of $l$, so define the token-normalized quantity $V^{\mathrm{tok}}_\lambda:=l\operatorname{Cov}(Z_k)$; the claim is $V^{\mathrm{tok}}_\lambda=F+O(\|B\|)$, which is what the sandwich needs because $\frac1{n_1}\operatorname{Cov}(Z_k)=\frac1{n_1l}V^{\mathrm{tok}}_\lambda=\frac1nV^{\mathrm{tok}}_\lambda$. Split the per-token score at $\theta^*_{\mathrm{post}}$ into a clean part $(\mu(A_{\theta^*_{\mathrm{post}}}\psi_i)-\phi(x_i))\psi_i^\top$ and a teacher part $\frac\lambda T\big(\mu(\tfrac{A_{\theta^*_{\mathrm{post}}}\psi_i}{T})-\mu(\tfrac{A_{\theta^*_{\mathrm{pre}}}\psi_i}{T})\big)\psi_i^\top$. Only the clean part is a martingale difference for the prefix filtration: conditionally on $\vx_{<i}$ the token $x_i$ has mean embedding $\mu(A_{\theta^*_{\mathrm{post}}}\psi_i)$, so clean parts at different positions of the same document are uncorrelated and their covariances add, contributing $\frac1{l^2}\sum_i\mathbb E[\Sigma(A_{\theta^*_{\mathrm{post}}}\psi_i)\otimes\psi_i\psi_i^\top]=\frac Fl$ to $\operatorname{Cov}(Z_k)$, hence $F$ to $V^{\mathrm{tok}}_\lambda$, with $F$ now read as an average over positions. The teacher part is \emph{not} a martingale difference: it is a deterministic function of $\psi_i$ whose conditional mean is generally nonzero, and its within-document correlations do not cancel. They are harmless only because the whole term is $O(\|B\|)$, so it adds $O(\|B\|^2)$ to $\operatorname{Cov}(Z_k)$, and the cross term with the clean part adds $O(\|B\|)$ there, only pairs $j>i$ surviving since $r_T(\psi_j)$ for $j\le i$ is measurable with respect to the prefix at $i$. In $V^{\mathrm{tok}}_\lambda$ these become $O(l\|B\|^2)$ and $O(l\|B\|)$, both absorbed into $V^{\mathrm{tok}}_\lambda=F+O(\|B\|)$ at fixed $l$. We do not claim a version with $l$ growing with $n_1$: there the within-document correlations of the teacher term enter a long-run variance, and a martingale central limit theorem would in any case not supply by itself the second-moment remainder that \eqref{eq:loglinear-risk} uses. No repetition of contexts is assumed anywhere.

\end{document}